\documentclass[preprint,12pt]{elsarticle}
\usepackage[top=1in,bottom=1in,left=1in,right=1in]{geometry}

\usepackage{amssymb}
\newcommand{\Tr}{\mbox{Tr}}
\usepackage{amsthm, amsmath}

\usepackage{txfonts}
\usepackage{setspace}
\usepackage{xcolor}
\usepackage{caption}
\usepackage{subcaption}
\usepackage{float}
\usepackage[pdfborder={0 0 0},colorlinks,allcolors=blue]{hyperref}
\theoremstyle{definition}
\usepackage{comment}
\usepackage{xcolor, soul}
\usepackage{diagbox}
\usepackage{multirow}

\definecolor{fgwhite}{rgb}{1,1,1}     
\definecolor{fgred}{rgb}{0.8,0,0}     
\definecolor{fgorange}{rgb}{0.93,0.53,0.18}     
\definecolor{fgpurple}{rgb}{0.55,0.1,0.6}     
\definecolor{fggreen}{rgb}{0,0.5,0}     
\definecolor{bggreen}{rgb}{0.8,1,0.8}     
\definecolor{fgblue}{rgb}{0,0,0.7}     
\definecolor{bgblue}{rgb}{0.9,0.9,1}     
\definecolor{fgclay}{rgb}{0.51,0.25,0.04}     
\definecolor{bggreen}{rgb}{0.8,1,0.8}     

\allowdisplaybreaks

\newcommand{\resub}[1]{{\color{black} #1}}
\newcommand{\rev}[1]{{\color{black} #1}}

\journal{Machine Learning: Science and Technology}

\begin{document}

\begin{frontmatter}

\title{Machine learning surrogate models of many-body dispersion interactions in polymer melts}

\author[inst1]{Zhaoxiang Shen}
\author[inst2]{Ra\'ul I. Sosa}
\author[inst1,inst3]{Jakub Lengiewicz}
\author[inst2]{Alexandre Tkatchenko}
\author[inst1]{St\'ephane P.A. Bordas\corref{cor1}}\ead{stephane.bordas@uni.lu}
\cortext[cor1]{Corresponding author}

\affiliation[inst1]{organization={Department of Engineering; Faculty of Science, Technology and Medicine; University of Luxembourg},
            city={Esch-sur-Alzette},
            country={Luxembourg}}

\affiliation[inst2]{organization={Department of Physics and Materials Science, University of Luxembourg},
            city={Luxembourg City},
            country={Luxembourg}}

\affiliation[inst3]{organization={Institute of Fundamental Technological Research, Polish Academy of Sciences},
            city={Warsaw},
            country={Poland}}

\begin{abstract}
Accurate prediction of many-body dispersion (MBD) interactions is essential for understanding the van der Waals forces that govern the behavior of many complex molecular systems. However, the high computational cost of MBD calculations limits their direct application in large-scale simulations. In this work, we introduce a machine learning surrogate model specifically designed to predict MBD forces in polymer melts, a system that demands accurate MBD description and offers structural advantages for machine learning approaches. Our model is based on a trimmed SchNet architecture that selectively retains the most relevant atomic connections and incorporates trainable radial basis functions for geometric encoding. We validate our surrogate model on datasets from polyethylene, polypropylene, and polyvinyl chloride melts, demonstrating high predictive accuracy and robust generalization across diverse polymer systems. In addition, the model captures key physical features, such as the characteristic decay behavior of MBD interactions, providing valuable insights for optimizing cutoff strategies. Characterized by high computational efficiency, our surrogate model enables practical incorporation of MBD effects into large-scale molecular simulations.
\end{abstract}


\end{frontmatter}

\section{Introduction}
\label{sec:Introduction}
Van der Waals (vdW) dispersion interactions play a critical role in diverse physical and chemical phenomena, from layered material cohesion \cite{liu2019van} and protein folding \cite{ProteinFolding} to gecko adhesion \cite{gecko}. Accurately capturing vdW dispersion forces is essential for understanding and predicting the properties of numerous engineering and biological systems. Well established pairwise (PW) models, such as Lennard-Jones (LJ) potential, Grimme's dispersion corrections~\cite{Grimme-ChemRev}, or the Tkatchenko–Scheffler (TS) method \cite{TS}, are widely used due to their simplicity and computational efficiency. While effective for certain systems, they inherently neglect the quantum many-body nature of vdW dispersion interactions, often leading to discrepancies with experimental observations \cite{Supramolecular,OrganicMolecularMaterials,Polymorphism}. For this reason, the many-body dispersion (MBD) method \cite{PhysRevLett.108.236402,doi:10.1063/1.4789814} becomes more preferable because it offers a more sophisticated and accurate approach by explicitly capturing collective long-range electron correlations. Numerous studies have shown that MBD surpasses PW models in reproducing experimental or nearly exact quantum-mechanical results, achieving superior accuracy in diverse scenarios, including thin-layer delamination \cite{Hauseux}, modeling of supramolecular systems \cite{Supramolecular}, and polymorphs prediction for molecular crystals \cite{polymorphs}.

The major obstacle of using MBD is its computational complexity, which scales as $O(N^3)$ with the number of atoms, $N$, thus severely limits its application to large-scale simulations. Recent efforts have attempted to accelerate MBD through physically motivated approximations and algorithmic optimizations of the analytical formulation \cite{stochastic_MBD,atom-wise_MBD}, but these approaches still require further refinement to achieve practical applicability in large-scale simulations. An alternative and more common approach is to develop a surrogate model that offers a fast yet accurate approximation, addressing the problem from the efficiency side while gradually improving accuracy. In molecular simulations, this is typically referred to as a force field (FF) model, which aims to approximate the high-fidelity energy and forces of molecular systems using functional forms based on simple geometric descriptors, such as interatomic distances, bond angles, and torsions \cite{AMBER,GROMOS05,OPLS,CHARMM-FF}. 

While classical FF models are computationally efficient, they suffer from limited accuracy due to their inadequate description of polarization and many-body interactions. To overcome these drawbacks, machine learning force fields (MLFFs) \cite{MLFF-2020,MLFF-2021-1,MLFF-2021-2, MLFF-2021-3,MLFF-book,PEIVASTE2025119419_ML_review} have emerged as promising solutions, offering quantum-level accuracy with competitive computational efficiency. MLFFs, which include a wide range of models from kernel-based methods to artificial neural networks (e.g., sGDML \cite{sGDML}, SchNet \cite{SchNet, SchNet_JCP}, PhysNet \cite{PhysNet}, GEMS \cite{GEMS}, among others \cite{GAP,SOAP,SOAP_ML,LODE,LODE-general,MLFF-phosphorus,FCHL,NN-PES-2007,NN-symmety-2011, So3krates_2022, SO3KRATES_2024, NequIP, MACE}), aim to learn the statistical relationship between chemical structures and the corresponding potential energy as well as atomic forces. Despite their advantages, many MLFFs are primarily trained for short-distance interactions and continue to rely on classical PW models for vdW interactions \cite{MLIP-LR-Perspective}. Some approaches, such as SpookyNet \cite{spookynet} and SO3LR \cite{SO3LR}, incorporate MBD-corrected reference data into their training while explicitly embedding parametric PW vdW functions within their architectures. These approaches, however, remain essentially parametric fittings to MBD and are limited in capturing many-body vdW effects beyond intramolecular interactions. To the
best of our knowledge, there is currently no MLFF model explicitly designed as an MBD surrogate capable of capturing many-body vdW effects in large-scale systems.

The objective in this work is to develop an ML surrogate model for MBD interactions that is specifically designed for \emph{large-scale systems}. Here, we focus on a particular class of molecular systems, the polymer melts, which can benefit from an accurate MBD description of vdW interactions, and which also requires a fast model for large-scale molecular simulations. This choice of systems follows the conclusions made in our previous work \cite{DFTB+MBD}, where we have identified classes of systems where MBD effects are most pronounced, namely, extended systems with flexible degrees of freedom whose physical properties are predominantly governed by vdW interactions, such as polymer crystals, polymer melts, and proteins. Among these, polymer melts emerge as the ideal initial target for our surrogate modeling. Polymer melts \cite{polymer_book_physics,Melt_Processes} are characterized by their disordered amorphous arrangement of polymer chains, where the chains exhibit significant mobility and fluid-like behavior while maintaining their high-molecular-weight structure. Accurately capturing vdW interactions is essential for predicting the mechanical and thermodynamic properties of polymer melts, as vdW forces directly influence key behaviors such as chain mobility, density, and viscosity \cite{polymer_book_properties,polymer_book_prediction}, while the desired molecular simulations of polymer melts often involve more than 100k atoms \cite{MD_PE_deformation,MD_PE_shear,MD_PE_crystal_Yamamoto,MD_PE_crystal_Theodorou}. Despite extensive efforts that have been made to enable such large-scale molecular simulations \cite{polymer_book_simulation, polymer_Hierarchical_Theodorou,polymer_roadmap,polymer_scales,polymer_ML}, incorporating MBD into these simulations still requires a dedicated ML architecture to provide an efficient surrogate model.

In this paper, we introduce a trimmed SchNet architecture that is specifically tailored for MBD force predictions in polymer melts. It is based on the SchNet architecture~\cite{SchNet, SchNet_JCP}, which we found an optimal choice for our problem due to its simplicity and superior adaptability (see also a more detailed discussion in Section~\ref{sec:model_requirements}). SchNet has demonstrated strong performance in molecular energy and force predictions, which is attributed to its radial basis function (rbf) encoding of structural information and continuous-filter convolutions. It remains scalable with respect to system size and ensures rotational invariance for energy prediction and rotational equivariance for force prediction. In our adaptation, we reduce the dimension and depth of the model while trimming the connection graph to retain only the most relevant interactions, which are then encoded using a reduced number of trainable rbfs. In addition, inspired by the repetitive structure of polymer chains, we introduce a unit-specific batching strategy that enhances training convergence. To validate the predictive capacity of the model, we apply it to three types of polymer melts: polyethylene (PE), polypropylene (PP), and polyvinyl chloride (PVC) and further assess its generalization capacity and robustness on mixed datasets and advanced MBD variants. A preliminary study is also conducted to evaluate temperature effects on the model’s performance. Furthermore, we demonstrate the physical interpretability of the model through analysis of the Hessian and discuss practical implementation of the model in molecular dynamics (MD) simulations together with a preliminary stability test, showcasing its applicability to large-scale systems. To increase the impact of this work, we open-source our codes and dataset, which are available in our repository \cite{trimmedSchNetRepo}. 

The remainder of this paper is structured as follows: Section~\ref{sec:methods} introduces the formalism of MBD and outlines the requirements for MBD surrogate modeling in polymer melts. It then details the proposed trimmed SchNet architecture, describes the data generation process, and provides implementation details for training. In Section~\ref{sec:results}, we evaluate the accuracy,  generalization capability, and robustness of the model through multiple examples, including three types of polymer melts, and analyze the key components of its architecture. In addition, we perform an analysis of the Hessian in this section and discuss the implementation of the model in MD simulations. Finally, Section~\ref{sec:conclusion} summarizes the work and discusses potential directions for future research.

\section{Methods}
\label{sec:methods}
In this section, we will first introduce the intricate formalism of MBD in comparison to the classical PW model, see Section~\ref{sec:mbd}. Then, in Section~\ref{sec:model_requirements}, we will outline the requirements for an MBD surrogate, which motivate the choice of a SchNet-based architecture. The proposed trimmed SchNet model, along with details on the trimmed connections, will be elaborated in Section~\ref{sec:SchNet}. In Section~\ref{sec:data}, we will specify the selected polymer melt systems and describe the corresponding data generation process. Finally, Section~\ref{sec:implementation} will present the training setup and evaluation metric.

\subsection{Many-body dispersion method}
\label{sec:mbd}
Van der Waals (vdW) dispersion, also know as London dispersion, is a ubiquitous attractive interaction between atoms and molecules. It originates from the temporary fluctuations in the mean-field electron density, leading to instantaneous dipoles that then interact with other dipoles induced by them. In this work, we focus exclusively on the dispersion interaction, which corresponds to the attractive part of the conventional vdW model. Traditional vdW modeling approaches rely on PW additive methods, where the dispersion energy $E^{\text{disp,PW}}$ for an $N$-atom molecular system is calculated as a simple sum of pairwise atomic interactions:
\begin{equation}
E^{\text{disp,PW}} = -\sum_{j>i}^{N}f^\text{damp} \dfrac{C_{6,ij}}{r_{ij}^{6}},
\label{eq:E_pw}
\end{equation}
where $r_{ij}=\|\boldsymbol{r}_{i}-\boldsymbol{r}_{j}\|$ is the interatomic distance between atoms $i$ and $j$, and $\boldsymbol{r}$ represents the atomic position. The material parameters $C_{6,ij}$ are determined experimentally or through high-fidelity numerical calculations. A damping function $f^\text{damp}$ is employed to mitigate non-physical singularities at short distances, while the function value converges to 1 at large distances. The interatomic force of a given atom $i$ in the system is obtained by taking the negative gradient of the energy with respect to its atomic position $\boldsymbol{r}_{i}$: 
\begin{equation}
\boldsymbol{F}_{\boldsymbol{r}_i}^{\text{disp,PW}}=-\nabla_{\boldsymbol{r}_i}E^{\text{disp,PW}}.
\end{equation}

However, the vdW dispersion interaction is inherently many-body in nature, as it arises from correlated electron fluctuations. The many-body dispersion (MBD) method \cite{PhysRevLett.108.236402,doi:10.1063/1.4789814} provides an efficient quantum-mechanical treatment of this phenomenon, offering a more accurate description compared to PW models. The MBD energy reads:
\begin{equation}
E^{\text{disp,MBD}} = \dfrac{1}{2} \sum_{l=1}^{3N} \sqrt{\lambda_l} - \dfrac{3}{2} \sum_{i=1}^{N} \omega_i, 
\label{eq:E_mbd}
\end{equation}
where $\lambda_l$ is the eigenvalue of a $3N\times3N$ matrix $\boldsymbol{C}_{}^{\text{MBD}}$ that couples each pair of atoms (see Eq.~\eqref{eq:Cmat}), and $\omega_i$ is the atomic characteristic frequency. To enhance readability, we present the detailed methodology and formalism of MBD in \ref{app_sec:mbd}. Taking the negative gradient of the energy with respect to the atomic position, after linear algebra transformations, the MBD force acting on atom $i$ can be expressed as: 
\begin{equation}
\boldsymbol{F}_{\boldsymbol{r}_i}^{\rm disp,MBD}=-\frac{1}{4} \Tr \left[ \boldsymbol{\Lambda}^{-1/2}\, \boldsymbol{S}^\text{T} \left(\nabla_{\boldsymbol{r}_i}\boldsymbol{C}^{\text{MBD}}\right)\boldsymbol{S}\right],\\
\label{eq:F_mbd}
\end{equation}
where $\boldsymbol{\Lambda}_{lq}=\lambda_l\delta_{lq}$ is obtained from the diagonalization of $\boldsymbol{C}^{\text{MBD}}$ via a suitable SO(3$N$) rotation matrix $\boldsymbol{S}$ that satisfies $\boldsymbol{\Lambda}=\boldsymbol{S}^\text{T}\boldsymbol{C}^{\text{MBD}}\boldsymbol{S}$. Summing over the eigenmodes reflects the collective contributions from the whole system, qualitatively differing from the pairwise force splitting in simpler PW models. 

As discussed in \cite{DFTB+MBD}, the correlated MBD model is fundamental for capturing cooperative motions that can significantly influence the dynamics of systems such as polymer melts. However, it involves the diagonalization operation for MBD forces as expressed by Eq.~\eqref{eq:F_mbd}, which introduces a computational complexity of $O(N^3)$, along with substantial memory demands compared to the additive PW formula. This makes direct application of MBD infeasible for large-scale polymer melt simulations, which can exceed 100k atoms. To overcome this challenge, an ML surrogate model with a tailored modeling strategy is required. In the following section, we analyze this modeling challenge from a computational perspective and demonstrate the requirements for an effective MBD surrogate for polymer melts, leading to the development of the proposed model architecture, as detailed in Section~\ref{sec:SchNet}.

\subsection{Surrogate modeling requirements for MBD in polymer melts}
\label{sec:model_requirements}

\begin{figure}[h]
 \centering
\subfloat[Unfolded unit cell.]{\includegraphics[trim = 10mm 60mm 10mm 60mm, clip=true,width=0.33\textwidth]{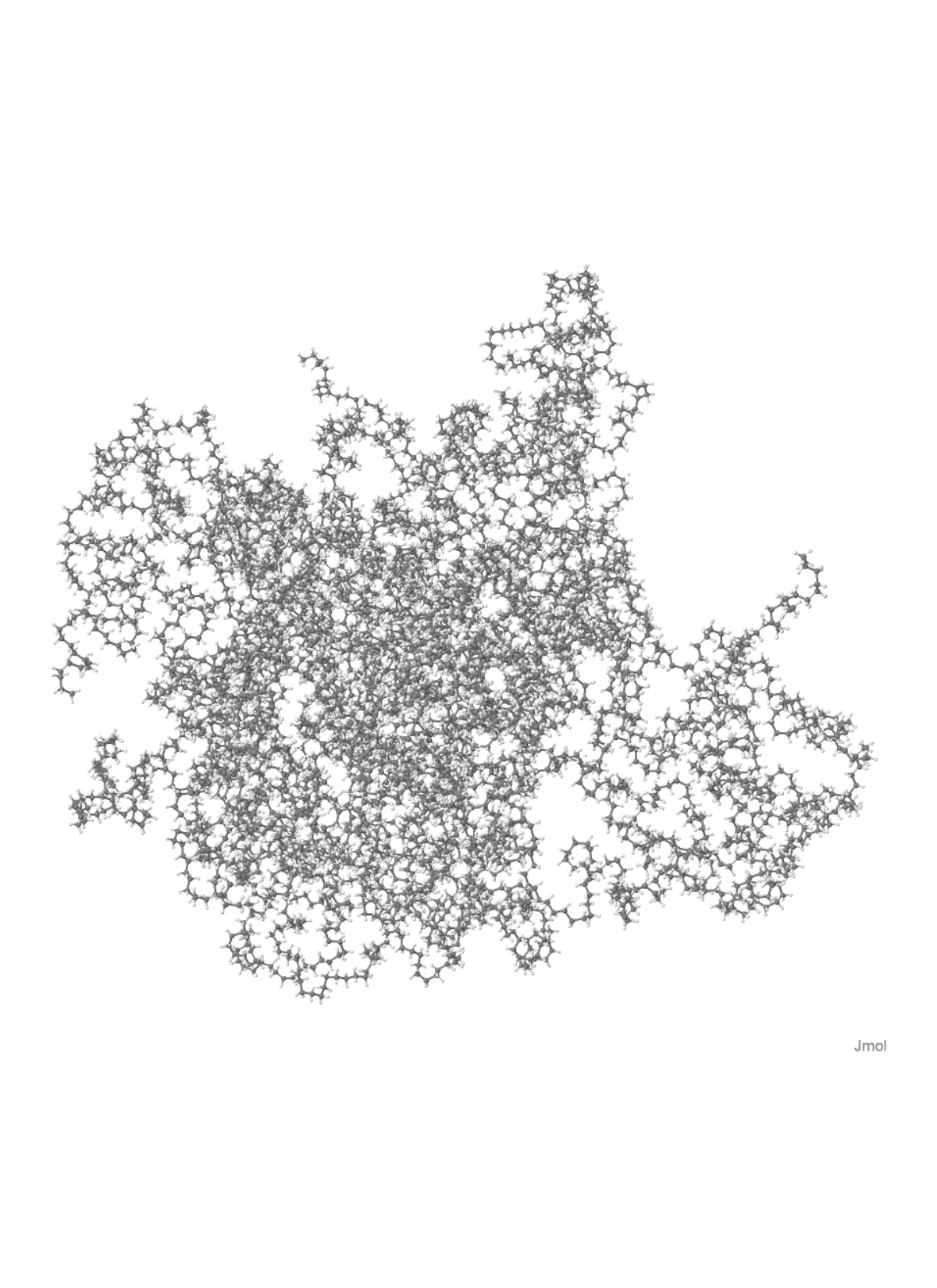}}
\subfloat[Folded unit cell.]{\includegraphics[trim = 10mm 60mm 10mm 60mm, clip=true,width=0.33\textwidth]{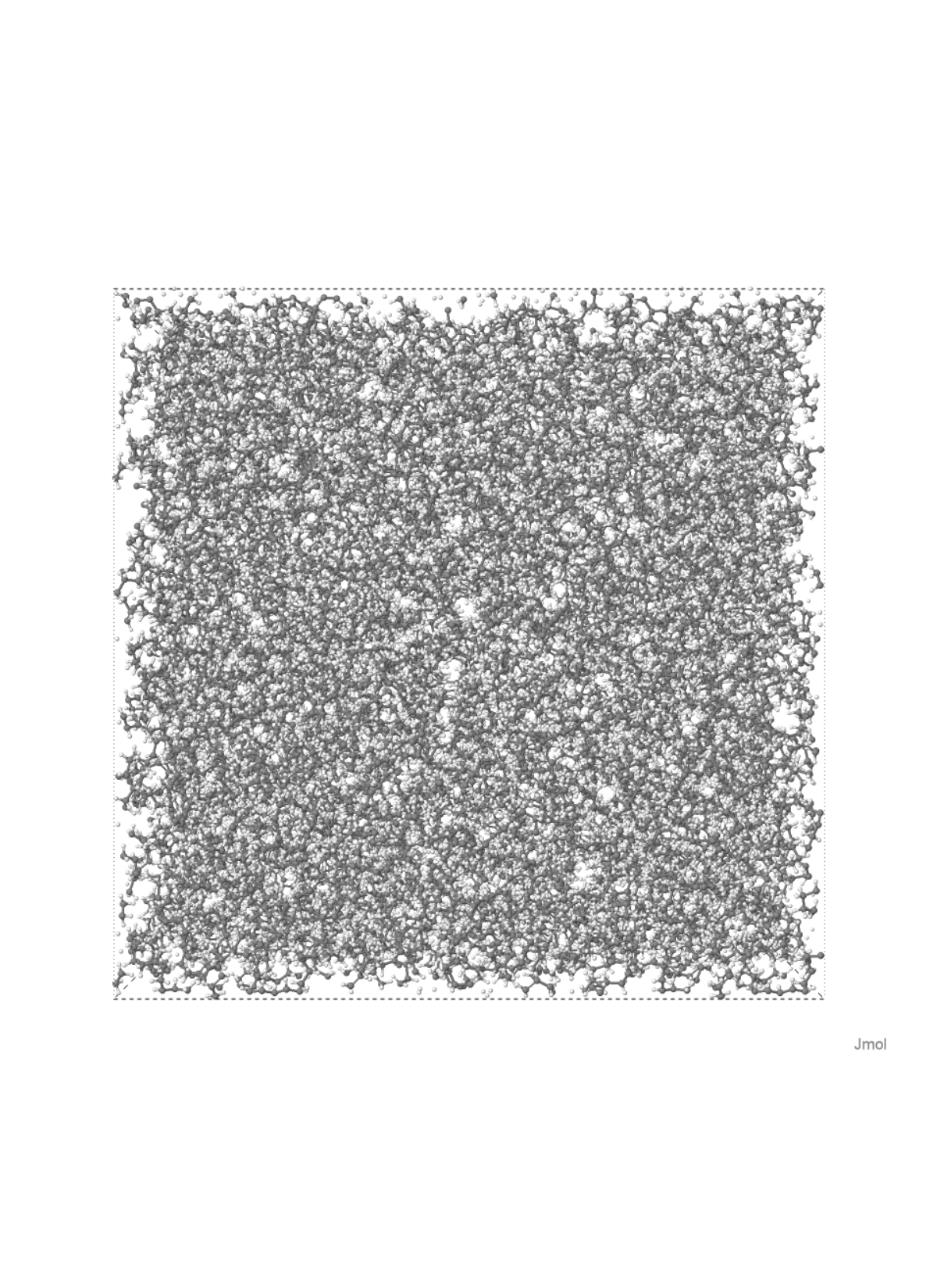}}
\subfloat[Cluster.]{\includegraphics[trim = 10mm 60mm 10mm 60mm, clip=true,width=0.33\textwidth]{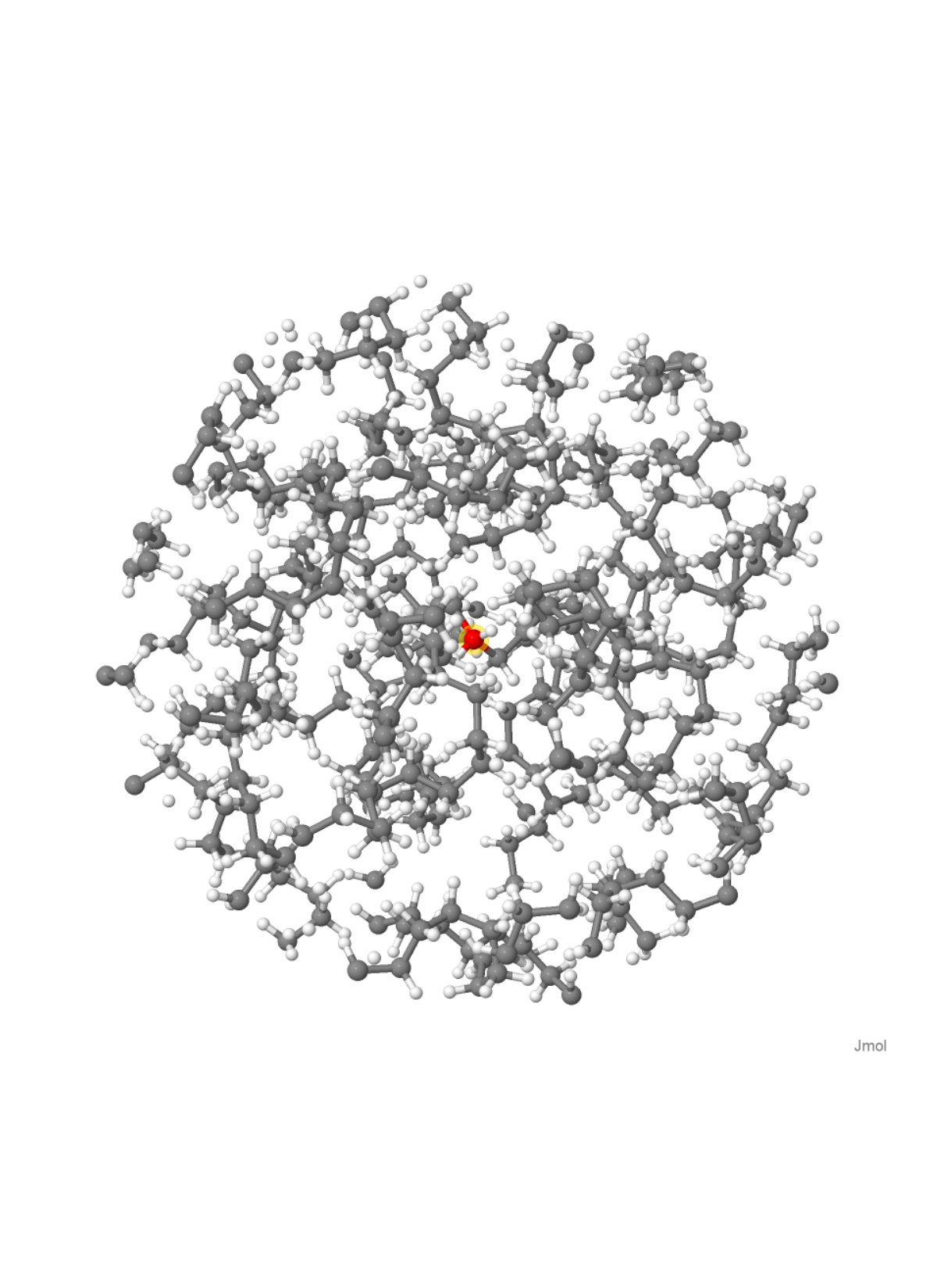}\label{fig:PE_cluster}}
    \caption{A PE melt structure. (a) is the original unfolded configuration of the unit cell. (b) is the folded unit cell according to the periodic boundary condition. (c) shows an atomic cluster with 1000 atoms cut off from (b), in which the center atom is marked as red.} \label{fig:PE}
\end{figure}

An effective surrogate model should be informed by both the numerical properties of the underlying physical model and the structural characteristics of the target system, in this case, MBD and polymer melts. As a reference, Fig.~\ref{fig:PE} presents a representative polyethylene (PE) melt structure. The dense and periodic system consists of a finite number of polymer chains within the unit cell, with chain lengths ranging from a few hundred monomers to millions depending on the molecular weight.

Even with conventional PW formulas, such as Eq.~\eqref{eq:E_pw}, computational feasibility is achieved by leveraging the decay properties of vdW dispersion to impose a cutoff range of interactions for each atom. This ensures that the overall computational complexity remains $O(N \times N_\text{cut})$, where $N_\text{cut}$ is the number of atoms within the cutoff neighborhood. To maintain efficiency, the MBD model must adopt a similar strategy. As illustrated in (a)-(c) of Fig.~\ref{fig:PE}, MBD calculations are performed on the nearly spherical atomic clusters extracted from the full folded system, a strategy directly employed in the proposed surrogate modeling approach. Specifically, the surrogate model processes an atomic cluster as input and outputs the MBD force acting on its center atom, which is repeated for all atoms in the system. Moreover, accurately predicting MBD forces requires a significantly larger cutoff range compared to PW models, due to the inherently long-range nature of many-body correlations, as discussed in \cite{MBD_Polymer_SOSA,Hauseux,wavelike_science_2016}. This presents an additional challenge, as $N_\text{cut}$ scales cubically with the cutoff distance. On the positive side, the structural characteristics of polymer melts introduce notable regularity in the cutoff region. For a fixed $N_\text{cut}$, the extracted atomic clusters closely resemble spheres with similar radii. This structural uniformity benefits statistical ML surrogate modeling, as the limited number of atom types enables the model to focus primarily on encoding local geometric details rather than learning higher-level chemical features.

Taking these considerations into account, we conclude that an optimal architecture for our problem, beyond fundamental requirements such as rotational equivariance for force predictions, should exhibit the following properties:
\begin{itemize}
    \item Lightweight design to enable efficient repetitive and parallel execution in large-scale simulations.
    \item Decomposability to allow per-atom force contributions, aligning with the center-focused nature of the problem.
    \item Strong capability to encode local geometric features while effectively capturing long-range interactions.
\end{itemize}
\rev{Several well-established MLFF models, such as SchNet, NequIP \cite{NequIP}, MACE \cite{MACE}, SpookyNet \cite{spookynet}, and SO3LR \cite{SO3LR}, satisfy these requirements and demonstrate strong general-purpose performance. However, many of these architectures are advanced derivatives of one another, designed for broad chemical spaces and enriched prediction targets, often incorporating higher-order features together with multiple layers of physically motivated interaction modules (e.g., attention-based modules \cite{attention}). While powerful, these additions introduce substantial computational overhead. In our case, the task is restricted to predicting MBD forces, a non-bonded interaction with a smoother landscape than the full interatomic physics, and is formulated for a single target atom per cutoff cluster. This narrower scope does not benefit from the additional complexity of such general-purpose models. Adapting them would likely involve removing most of their specialized components, effectively converging towards a model similar to SchNet. For this reason, we adopt a SchNet-based design, starting from a minimal yet robust baseline that avoids the overhead of unused functionality while retaining the required accuracy. This deliberate trade-off of generality for scalability makes the model well matched to our task. In the following section, we present our proposed trimmed SchNet and the dedicated modifications tailored to the MBD surrogate problem.}


\subsection{Trimmed SchNet}
\label{sec:SchNet}
We now present the architecture of the trimmed SchNet, designed for surrogate modeling of MBD forces in polymer melts. The section begins with an overview of the proposed architecture, followed by a detailed discussion of key modifications to the original SchNet, including trimmed connections (Section~\ref{sec:trimmed_connections}) and trainable rbf encoding (Section~\ref{sec:rbf}). Finally, Section~\ref{sec:SchNet_force} presents the modified model output and the force-based loss function.

\begin{figure}[h]
\centering
\includegraphics[trim = 5mm 35mm 5mm 35mm, clip=true,width=1\textwidth]{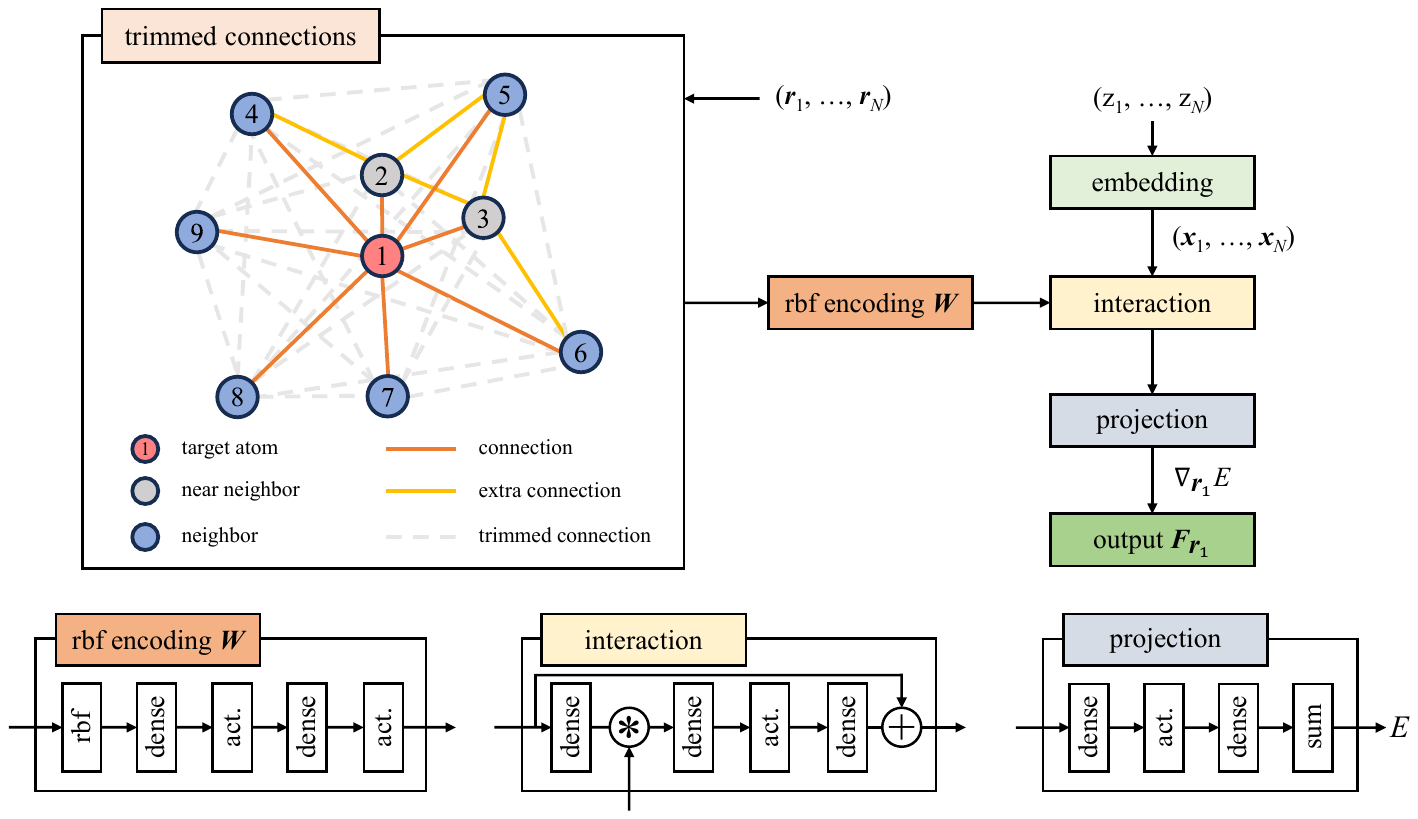}
    \caption{Illustration of the trimmed SchNet. The trimmed connections graph is depicted for a toy system that consists of $N=9$ atoms, where only connections to the central target atom (red solid lines) and a few extra connections to its two nearest neighbors (yellow solid lines) are retained, while all other peripheral connections (gray dashed lines) are trimmed. The detailed architectures of rbf encoding, interaction, and projection are shown in the bottom.}
     \label{fig:SchNet_diagram}
\end{figure}

\subsubsection{Architecture overview}
The trimmed SchNet model processes an atomic system by taking atomic positions and atom types as inputs, encoding their structural information, and predicting the MBD force of the center atom. Given an atomic system with $N$ atoms, the model takes the atomic positions $\boldsymbol{R}=\left(\boldsymbol{r}_1,...,\boldsymbol{r}_N\right)$ and the corresponding atom types $\boldsymbol{Z}=\left(\text{z}_1,...,\text{z}_N\right)$ as inputs. For expressiveness, each atom is represented by a $P$-dimensional feature embedding, $\boldsymbol{x}_i^l \in \mathbb{R}^P$, where $\boldsymbol{x}_i^l$ is the feature vector at layer $l$. These embeddings are initialized based on the atom types $\boldsymbol{Z}$ and iteratively updated through interaction layers.

Fig.~\ref{fig:SchNet_diagram} illustrates the architecture of the trimmed SchNet. The model follows the fundamental structure of the original SchNet \cite{SchNet,SchNet_JCP}, where atomic interactions are modeled through learned feature representations. It encodes structural information based on specified atomic connections, processes interactions via continuous-filter convolutions, and projects the learned representations into a scalar energy value. The force output is then obtained as the gradient of the energy with respect to atomic positions.

While preserving SchNet’s core structure, the proposed architecture introduces several modifications tailored to the MBD surrogate problem. These include trimming interaction connections, incorporating trainable rbf encoding, using a single interaction block, and reducing the network width. These adaptations simplify the model while maintaining its functionality, enabling efficient implementations for large-scale molecular simulations. 

\subsubsection{Interactions by trimmed connections}
\label{sec:trimmed_connections}
Formally, the atomic feature updates of atom $i$ in SchNet are computed by an interaction block with so called continuous-filter (cf) convolution:
\begin{equation}
\boldsymbol{x}_i^{l+1} = \boldsymbol{x}_i^l + \sum_{j \neq i}^N \boldsymbol{W}^{l}(\boldsymbol{r}_i-\boldsymbol{r}_j) * \boldsymbol{x}_j^l, 
\end{equation}
where $\boldsymbol{W}$ is the filter-generating function that encodes the atomic connections via radial basis functions, and $*$ denotes element-wise multiplication.

In the application to MBD force prediction for polymers, we only need to calculate for the center atom in a given cutoff range. To maintain a fixed input dimension for the neural network, the atomic cluster is cut off by a fixed number of atoms $N=N_\text{cut}$, while the position tuple $\boldsymbol{R}$ is ordered by the atomic distance to the center atom. In the original version, a full SchNet leverages all $\sum^{N_\text{cut}}_{n=2}(n-1)$ connections between atoms, which restricts the cutoff range in practice. However, our specific goal, focusing only on the center atom, implies that a reduction of required connections is potentially feasible. In this paper, we implement a trimmed SchNet that remains limited connections, as shown in connection graph in Fig.~\ref{fig:SchNet_diagram} \rev{(top left)}, which updates the atom feature as follows:
\begin{itemize}
    \item Keep the full convolution \rev{(orange lines)} to the atom $1$, i.e. the center atom \rev{(red circle)}:
    \begin{equation}
\boldsymbol{x}_1^{l+1} = \boldsymbol{x}_1^l + \sum_{j\neq 1}^{N_\text{cut}} \boldsymbol{W}(\boldsymbol{r}_1-\boldsymbol{r}_j) * \boldsymbol{x}_j^l.
    \end{equation}
    
    \item Add $N_\text{extra}$ connections \rev{(yellow lines)} for the $p$ nearest neighbors \rev{(gray circles)} of atom $1$:
    \begin{equation}
    \boldsymbol{x}_i^{l+1} = \boldsymbol{x}_i^l + \sum_{j\neq i}^{i+N_\text{extra}} \boldsymbol{W}(\boldsymbol{r}_i-\boldsymbol{r}_j) * \boldsymbol{x}_j^l, \ i=2,...,p+1.
    \end{equation}

    \item The rest of the atoms \rev{(blue circles)} only take reverse connections from the previous two updates:
    \begin{equation}
\boldsymbol{x}_i^{l+1} = \boldsymbol{x}_i^l + \sum_{j=1}^{p+1} \boldsymbol{W}(\boldsymbol{r}_i-\boldsymbol{r}_j) * \boldsymbol{x}_j^l, \ i=p+2,...,2+N_\text{extra}. 
    \end{equation}
    \begin{equation}
\boldsymbol{x}_i^{l+1} = \boldsymbol{x}_i^l + \sum_{j=i-N_\text{extra}}^{p+1} \boldsymbol{W}(\boldsymbol{r}_i-\boldsymbol{r}_j) * \boldsymbol{x}_j^l+ \boldsymbol{W}(\boldsymbol{r}_i-\boldsymbol{r}_1) * \boldsymbol{x}_1^l, \ i=3+N_\text{extra},...,p+1+N_\text{extra}. 
    \end{equation}    
    \begin{equation}
\boldsymbol{x}_i^{l+1} = \boldsymbol{x}_i^l + \boldsymbol{W}(\boldsymbol{r}_i-\boldsymbol{r}_1) * \boldsymbol{x}_1^l, \ i=p+2+N_\text{extra},...,N_\text{cut}. 
    \end{equation}
\end{itemize}
\rev{The trimmed connections, i.e., those disregarded from the original SchNet, are depicted as dashed gray lines.} Trimming these peripheral connections effectively reduces the computational cost of the model, while the remaining extra connections to the nearest neighbors enhance the capability of the trimmed SchNet. The results presented in Section~\ref{sec:extra_connnections} highlight the importance of these additional connections for achieving highly accurate fitting. Moreover, in practice, a SchNet model can incorporate multiple interaction blocks in sequence, each with separate rbf encodings, allowing it to capture more complex and hierarchical features as the model depth increases. In this work, we opt for a single interaction block to balance accuracy with computational efficiency.

\subsubsection{Rbf encoding}
\label{sec:rbf}
The filter-generating function $\boldsymbol{W}$ calculates the distance of input atoms $d_{ij}=\|r_i-r_j\|$, and then expands it by $N_\text{rbf}$ Gaussian radial basis functions (rbfs):
\begin{equation}
    \boldsymbol{e}_{k} = \text{exp}(-\gamma_k|d_{ij}-\mu_k|^2),
    \label{eq:rbf}
\end{equation}
where $\gamma_k$ is the rbf coefficient and $\mu_k$ is the center of an rbf, which forms the weighted encoding bases over the range of interest. In contrast to the original SchNet, we set both parameters as trainable to reduce the requirement of $N_\text{rbf}$, improving the convergence of training. The mechanism is that the tunable rbf centers tend to cluster to some critical distances over the cutoff range and are weighted accordingly by $\gamma_k$, therefore, encode the atomic structure more effectively. A numerical analysis for the trainable rbfs is shown in Section~\ref{sec:rbf_analysis}.   

\subsubsection{Force-based loss function}
\label{sec:SchNet_force}
The output of this trimmed SchNet is directly and only the force of the center atom, that is, atom 1. This is derived by taking the gradient of $E$ w.r.t. the atomic position:
\begin{equation}
    \boldsymbol{F}_{\boldsymbol{r}_{1}} = -\nabla_{\boldsymbol{r}_{1}} E,
    \label{eq:F_gradE}
\end{equation}
\resub{where $E$ here is only an energy-like latent quantity obtained after the projection layer of the network. Importantly, this quantity is not intended to represent a physically meaningful or conserved potential energy, as the model is trained on truncated atomic clusters extracted from a larger polymer melt system, for which a well-defined global energy does not exist.} Nevertheless, the rbf encoding introduced above ensures rotationally invariance of $E$, so by construction, the force defined by Eq.~\eqref{fig:SchNet_diagram} is rotationally equivariant.

With this \resub{force-only} output, our trimmed SchNet is trained with a reduced loss function compared to the one for the original SchNet. If we denote this neural network as $h: (\boldsymbol{R},\boldsymbol{Z})\rightarrow \boldsymbol{F}_{\boldsymbol{r}_{1}}$ and its trainable parameters as $\boldsymbol{\theta}$, the mean squared error loss function used in this paper can be formulated as:

\begin{equation}
    \mathcal{L}\left(\mathcal{D}_\text{train},\boldsymbol{\theta}\right)=\frac{1}{N_\text{train}}\sum_{i=1}^{N_\text{train}}\|h\left((\boldsymbol{R}_{i},\boldsymbol{Z}_{i}),\boldsymbol{\theta}\right)-\hat{\boldsymbol{F}}_{\boldsymbol{r}_{1}i}\|^2,
\end{equation}
where $\mathcal{D}_\text{train}=\{((\boldsymbol{R}_{1},\boldsymbol{Z}_{1}),\hat{\boldsymbol{F}}_{\boldsymbol{r}_{1}1}),...,((\boldsymbol{R}_{N_\text{train}},\boldsymbol{Z}_{N_\text{train}}),\hat{\boldsymbol{F}}_{\boldsymbol{r}_{1}N_\text{train}})\}$ is a given training dataset with $N_\text{train}$ data points, and $\hat{\boldsymbol{F}}_{\boldsymbol{r}_{1}}$ is the reference MBD force. This force loss requires the model to be at least twice differentiable for implementing gradient-based optimization methods. Therefore, SchNet uses a shifted softplus as the activation function: 
\begin{equation}
    \text{ssp}\left(x\right) = \ln{\left(e^x+1\right)} - \ln{2},
    \label{eq:ssp}
\end{equation}
which promises a smooth energy surface and improves the convergence by ensuring $\text{ssp}\left(0\right)=0$.

In principle, a SchNet model that solely outputs energy is simpler and requires only half the computational cost \cite{SchNet}. \resub{However, our goal is to develop a surrogate model for MBD forces that can be directly integrated into MD simulations. If the model predicts energies only, forces must be obtained at every MD step via an additional differentiation operation, either through a finite difference algorithm or through automatic differentiation (AD).} This extra step would inevitably introduce higher computational costs and pose additional challenges for deployment in external MD packages.

\subsection{Three polymer melt systems: description and data generation}
\label{sec:data}
\begin{figure}[h]
 \centering
\subfloat[PE.]{\includegraphics[trim = 40mm 95mm 40mm 95mm, clip=true,width=0.33\textwidth]{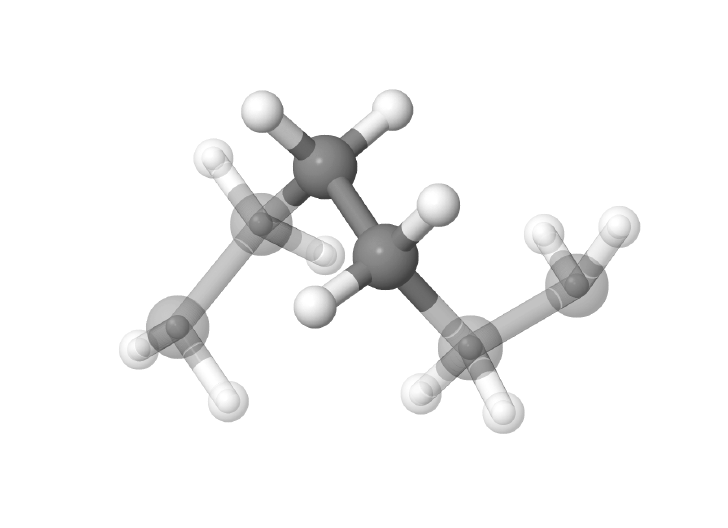}}
\subfloat[PP.]{\includegraphics[trim = 40mm 95mm 40mm 95mm, clip=true,width=0.33\textwidth]{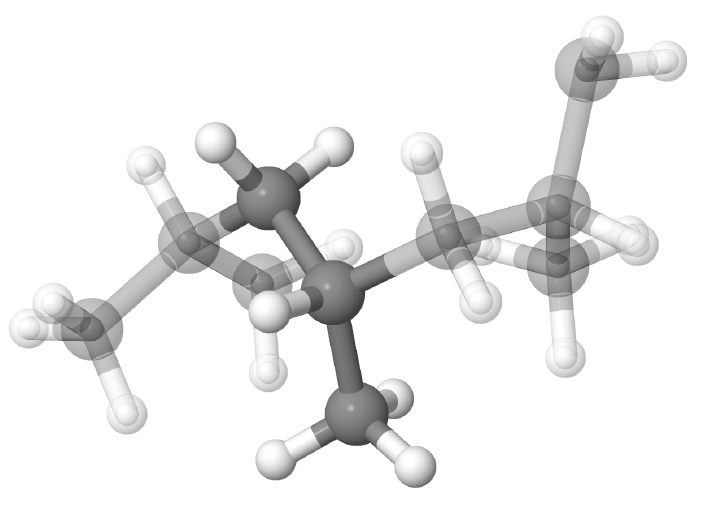}}
\subfloat[PVC.]{\includegraphics[trim = 40mm 95mm 40mm 95mm, clip=true,width=0.33\textwidth]{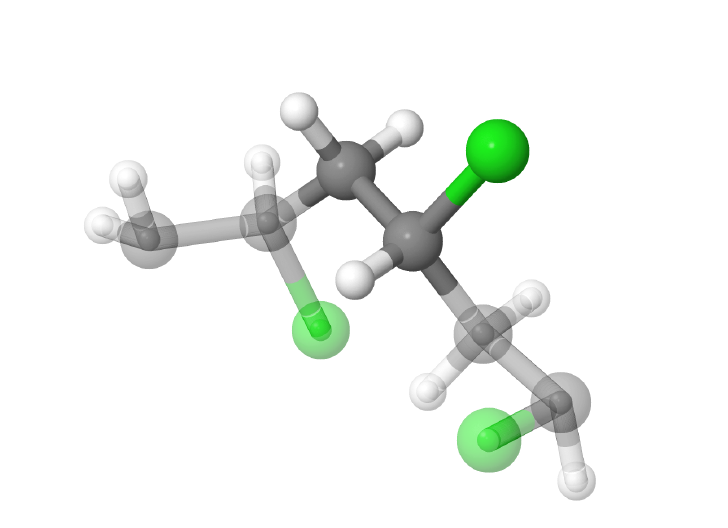}}
    \caption{Segments of the three types of polymer used in this work, with carbon atoms in dark gray, hydrogen atoms in white, and chlorine atoms in green. The monomer of each polymer (ethylene for PE, propylene for PP, and vinyl chloride for PVC) is highlighted in non-transparent shading.} \label{fig:monomer}
\end{figure}

To construct a dataset with diverse structural variability, three types of polymer melts, polyethylene (PE), polypropylene (PP), and polyvinyl chloride (PVC), are generated using the polymer builder in CHARMM-GUI \cite{CHARMM-GUI,CHARMM-GUI-polymer}. These polymers span from the simplest structure (PE) to more chemically complex systems: PP incorporates additional side-chain functional groups, while PVC features one more atom type, as presented in Fig.~\ref{fig:monomer}. 

For each polymer, multiple configurations with varying chain lengths are considered while maintaining a similar unit cell size across cases: $24\times100$, $12\times200$, $8\times300$, $6\times400$, and $4\times600$, where $n\times m$ denotes $n$ polymer chains, each with $m$ monomers. All of these systems are equilibrated using CHARMM-GUI at 300\,K to ensure uniform morphology within each polymer type.\footnote{PVC at 300\,K is below its glass transition temperature ($T_g \approx 350\,\text{K}$) and thus technically in a glassy state with very limited mobility. However, we broadly retain the term ``polymer melt'' throughout this study for simplicity and consistency. PVC is intentionally included to assess the model’s generalization across polymers with different structural and chemical characteristics.} In this setting, the dataset should naturally capture various levels of crystallinity across the three polymers, arising from their intrinsic molecular structures and thermal behaviors \cite{polymer_crystallization_2003,polymer_crystallization_2007}, which further enriches the dataset. For example, PE, with a glass transition temperature ($T_\text{g}$) of approximately 160\,K, can exhibit more crystalline regions at 300\,K due to the high chain mobility above its $T_\text{g}$ and its simple and highly symmetric structure for chain packing. In contrast, PVC, with a $T_\text{g}$ around 350\,K, remains largely amorphous at 300\,K as its polymer chains are relatively rigid below $T_\text{g}$ and its bulky chlorine side groups further hinder the regular chain packing required for crystallization. However, we acknowledge that the morphological effects, particularly crystallinity, require a more detailed investigation. Due to the limited number of time steps allowed by CHARMM-GUI, the polymer systems may not achieve sufficient crystallization by the online equilibrating procedure. In this paper, we present a preliminary study on this topic in \ref{sec:temperature}, while leaving a comprehensive exploration of crystallinity effects for future work.

For training the neural network, we collect 60k data points for PE, 72k for PP, and 60k for PVC, with $10\%$ of each reserved for validation. These data points are extracted equally from different configurations for each type of polymer. Each data point is a $\left(\left(\boldsymbol{R}, \boldsymbol{Z}\right),\hat{\boldsymbol{F}}_{\boldsymbol{r}_1}\right)$ tuple, representing an atomic cluster with $N_\text{cut}=1000$ atoms and the MBD force on its center atom (scaled by $10^3$ for numerical stability) generated by our open-source toolkit \cite{DFTB+MBD}. This number of $N_\text{cut}$ corresponds to a cutoff distance of approximately $14\,\text{\AA}$, significantly larger than the typical range of $6\text{-}9\,\text{\AA}$ used in classical PW methods due to the long-range nature of MBD. While $N_\text{cut}$ could be adjusted depending on the specific MBD variant and the density of the specific polymer melt, or reduced to accommodate stricter computational resource constraints, we select $N_\text{cut} = 1000$ as a general-purpose value for demonstration, ensuring applicability across all cases in this paper. Our cutoff strategy is consistent with a recently reported statistical analysis of MBD forces in similar PE melt systems \cite{MBD_Polymer_SOSA}. \resub{At this cutoff, the contributions from peripheral atoms decay smoothly towards negligible magnitudes, such that the outer region effectively serves as a buffer and changes in the cutoff neighborhood during MD lead to smooth force variations rather than discontinuous jumps, analogous to standard cutoff treatments in classical PW vdW models.} The selected clusters always include complete data points from the smallest repeating structural units in the polymers: methylene groups ($\text{C}\text{H}_2$) for PE, propylene monomers ($\text{C}_3\text{H}_6$) for PP, and vinyl chloride monomers ($\text{C}_2\text{H}_3\text{Cl}$) for PVC. This enables a special batching strategy to improve convergence of training as discussed in Section~\ref{sec:batching}. Regarding the edge atoms of polymer chains, although they are not directly involved in the training due to their minor representation in the structure, the model demonstrates reasonable predictive capability for them after being trained on our sufficiently extensive dataset.

\subsection{Implementation details}
\label{sec:implementation}
\subsubsection{Training setting}
\label{sec:training_setting}
\begin{figure}[h]
 \centering
\subfloat[PE.]{\includegraphics[trim = 30mm 90mm 40mm 90mm, clip=true,width=0.33\textwidth]{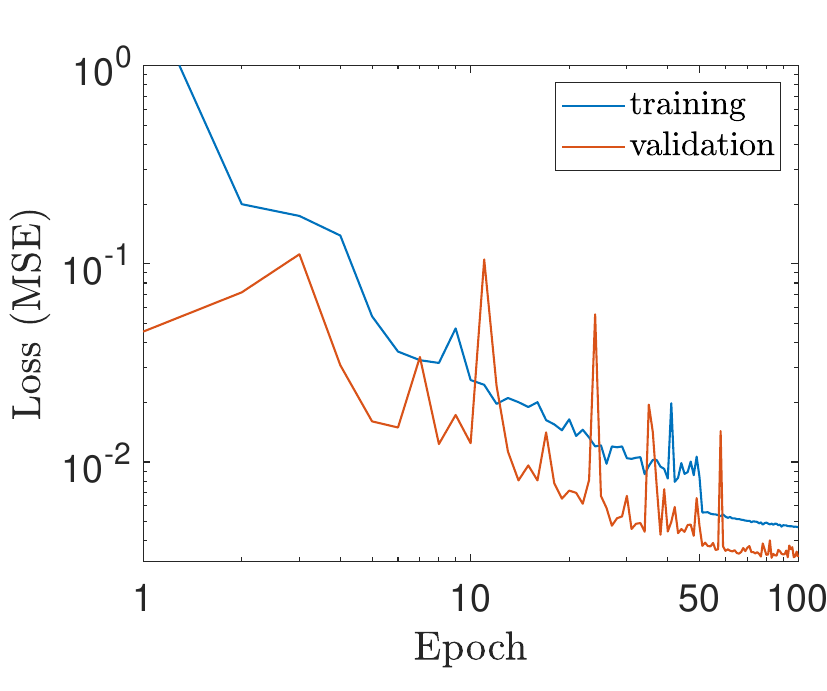}\label{fig:train_valid_PE}}
\subfloat[PP.]{\includegraphics[trim = 30mm 90mm 40mm 90mm, clip=true,width=0.33\textwidth]{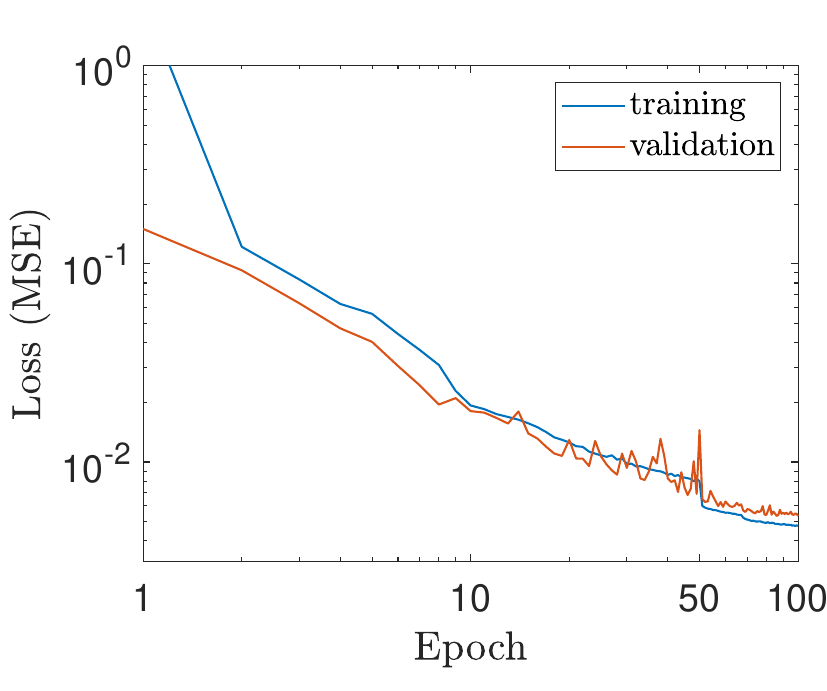}\label{fig:train_valid_PP}}
\subfloat[PVC.]{\includegraphics[trim = 30mm 90mm 40mm 90mm, clip=true,width=0.33\textwidth]{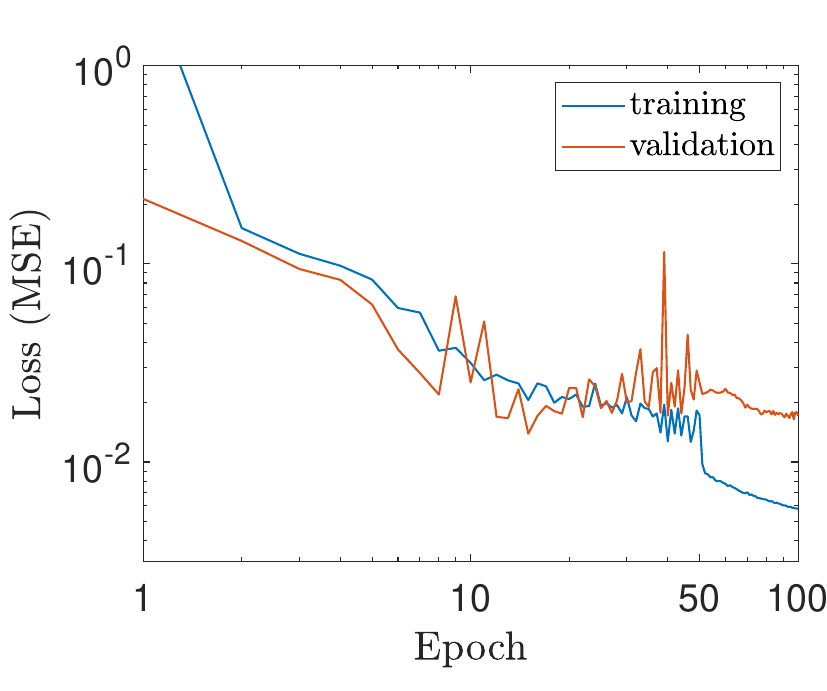}\label{fig:train_valid_PVC}}
    \caption{Training and validation convergence of the model on three polymer datasets, following the proposed training strategy.} \label{fig:train_valid}
\end{figure}

In the implementation, the embedding dimension is set to $P=32$, and is maintained throughout the rbf encoding block and the interaction block. The projection block at the end reduces the dimension to 16 and then to 1 by two dense layers. Again, we reduce $P$ by a factor of 2 compared to the original SchNet, which is tailored to our surrogate modeling problem. $p=2$ nearest neighbors are included for the extra connections with $N_\text{extra}=50$, and we use $N_\text{rbf}=100$ for the rbf encoding. The effects of these two settings are discussed in Section~\ref{sec:Model_analysis}. For each case in the following subsection, unless otherwise specified, the model is trained for 100 epochs with a batch size of 36. The learning rate is initialized at $10^{-3}$ and reduces to $10^{-4}$ after 50 epochs. AdamW \cite{adamW} is used as the optimizer with a weight decay of $0.004$. All the implementations in this work are carried out using TensorFlow and executed on the HPC facilities of the University of Luxembourg \cite{ULHPC}. The proposed training strategy is generally effective for the three polymers considered in this work, as shown in Fig.~\ref{fig:train_valid}. Given the robustness of the SchNet-based architecture, our focus is on understanding the model and its adaptation to the MBD surrogate problem. Fine-tuning hyperparameters and searching for the optimal training strategy for each dataset are beyond the scope of this paper.

\subsubsection{Evaluation metrics}
To evaluate the performance of the trained models, an additional test dataset consisting of 6k data points is collected for each polymer. Due to the uneven distribution of atom types within the polymers and the generally smaller average MBD force magnitudes observed on backbone atoms, e.g., carbon atoms in this study, the neural network naturally fits better for hydrogen and chlorine atoms. This trend is evident in the results shown later. Furthermore, the dataset inherently contains near-zero force components, arising from coincidental symmetries in the molecular geometry. Considering these factors, we adopt the mean absolute relative error (MARE) as the evaluation metric (in percentage) w.r.t mean force magnitude: 
\begin{equation}
    \mathcal{E}_\text{MARE}\left(\mathcal{D}_\text{test},\boldsymbol{\theta}\right) = \frac{1}{3N_\text{test}}\sum_{i=1}^{N_\text{test}}\sum_{j=1}^{3}\frac{|h\left((\boldsymbol{R}_{i},\boldsymbol{Z}_{i}),\boldsymbol{\theta}\right)_{j}-\hat{{F}}_{\boldsymbol{r}_{1}ij}|}{\tilde{{F}}_{\boldsymbol{r}_{1}}}\times100\%,
    \label{eq:e_MARE}
\end{equation}
where $\tilde{{F}}_{\boldsymbol{r}_{1}}=\frac{1}{3N_\text{test}}\sum_{i=1}^{N_\text{test}}\sum_{j=1}^{3}|\hat{{F}}_{\boldsymbol{r}_{1}ij}|$ is the mean force magnitude per degree of freedom. This approach ensures an unbiased assessment of the model's predictive capability. When evaluating for one specific type of atom, $\tilde{{F}}_{\boldsymbol{r}_{1}}$ will be the mean of only that kind. 

Later in this paper, we also occasionally use two other metrics to support analysis of model's performance:
\begin{equation}
    \mathcal{E}_\text{angle}\left(\mathcal{D}_{\text{test},i},\boldsymbol{\theta}\right)=G\left(\boldsymbol{h}\left((\boldsymbol{R}_{i},\boldsymbol{Z}_{i}),\boldsymbol{\theta}\right),\hat{\boldsymbol{F}}_{\boldsymbol{r}_{1}i}\right),
    \label{eq:e_angle}
\end{equation}
that measures the error of force angle ($^\circ$) of an atom with $G(\boldsymbol{u},\boldsymbol{v})=\arccos{\left(\frac{\boldsymbol{u}\cdot\boldsymbol{v}}{\|\boldsymbol{u}\| \|\boldsymbol{v}\|}\right)}\times\frac{180}{\pi}$, and
\begin{equation}
    \mathcal{E}_\text{ARE}\left(\mathcal{D}_{\text{test},i},\boldsymbol{\theta}\right)_j = \frac{|h\left((\boldsymbol{R}_{i},\boldsymbol{Z}_{i}),\boldsymbol{\theta}\right)_{j}-\hat{{F}}_{\boldsymbol{r}_{1}ij}|}{\tilde{{F}}_{\boldsymbol{r}_{1}}}\times100\%,
    \label{eq:e_ARE}
\end{equation}
that provides the force error per degree of freedom of an atom. 

\section{Results and discussions}
\label{sec:results}
In this section, we will demonstrate the predictive capability and robustness of the proposed trimmed SchNet model, enhanced by our novel modifications. First, in Section~\ref{sec:across_polymers}, we will present the model's performance on the three polymer melt systems, including mixed-polymer datasets. Then, in Section~\ref{sec:Model_analysis}, three key components of the model will be analyzed: trainable rbf encoding, specialized extra connections, and a dedicated unit-specific batching strategy. In Section~\ref{sec:advanced_MBD}, we will further examine the performance of the model for advanced MBD variants, which will be followed, in Section~\ref{sec:temperature}, by a preliminary study on temperature effects. A physically meaningful analysis of the model's Hessian will be presented in Section~\ref{sec:hessian}, and finally Section~\ref{sec:MD} will discuss the prospects of implementing our ML surrogate model in MD simulation environments.

\subsection{Model performance across polymers}
\label{sec:across_polymers}
The ability of the model to efficiently and accurately predict forces and generalize across different polymer systems is crucial for its applicability in large-scale molecular simulations. The trimmed SchNet model, with its lightweight architecture, enables fast inference, achieving computation times of 3\,ms/atom (batch size of 1) and 0.13\,ms/atom (batch size of 1000), depending on parallelization and memory constraints. This efficiency allows for extensive evaluations across different polymer systems. Furthermore, the practical execution of the trained model in MD simulations, where computational requirements are more complex and evolving, has also been proven to be efficient in Section~\ref{sec:MD}.

To assess its predictive accuracy and generalization capability, we trained the model on both single- and mixed-polymer datasets following the training setup described in Section~\ref{sec:training_setting}. The trained models were then tested on various polymer test datasets to evaluate their ability to transfer across polymer types and adapt to unseen structures. The test results are presented in Tab.~\ref{tab:mixed_polymer}.

\begin{table}[h]

    \caption{Performance ($\mathcal{E}_\text{MARE}$\,(\%)) of the model trained on single- and mixed-polymer datasets. \rev{Values in parentheses indicate standard deviations of $\mathcal{E}_\text{MARE}$\,(\%).}}
    
    \centering
    \resizebox{\textwidth}{!}{
    \begin{tabular}{|c|c|c|c|c|c|c|c|c|c|c|}
    \hline
        \multirow{2}{*}{\diagbox{Train}{Test}} & \multicolumn{3}{c|}{PE} & \multicolumn{3}{c|}{PP} & \multicolumn{4}{c|}{PVC} \\ \cline{2-11}
         & C & H & all & C & H & all & C & H & Cl & all \\ \hline
        \multirow{2}{*}{PE}  & \textbf{5.59} & \textbf{0.41}& \textbf{0.71}& 59.61 & 2.04 & 5.40 & - & - & - & - \\ 
        & \rev{(7.82)} & \rev{(0.57)} & \rev{(1.03)} & \rev{(80.49)} & \rev{(4.02)} & \rev{(10.03)} & - & - & - & - \\ \hline

        \multirow{2}{*}{PP}  & 15.15  & 0.46  & 1.30 & \textbf{6.38} & \textbf{0.49} & \textbf{0.83}  & - & - & - & - \\ 
        & \rev{(12.26)} & \rev{(1.63)} & \rev{(2.41)} & \rev{(6.90)} &  \rev{(0.45)} & \rev{(0.89)} & - & - & - & - \\ \hline

        \multirow{2}{*}{PVC}  & 43.13 & 3.53 & 5.78 & 49.81 & 3.41 & 6.12 & \textbf{3.42} & \textbf{0.60} & \textbf{1.00} & \textbf{0.98} \\
        & \rev{(28.63)} & \rev{(5.00)} & \rev{(6.52)} & \rev{(40.59)} & \rev{(2.61)} & \rev{(5.41)} & \rev{(3.40)} & \rev{(1.61)} & \rev{(0.87)} & \rev{(1.75)} \\ \hline

        \multirow{2}{*}{PE+PP} & 6.54 & 0.44 & 0.79 & 8.03 & 0.56 & 1.00 & - & - & - & - \\
        & \rev{(8.57)} & \rev{(0.58)} & \rev{(1.10)} & \rev{(8.14)} & \rev{(0.51)} & \rev{(1.05)} & - & - & - & - \\ \hline
        
        \multirow{2}{*}{PE+PVC} & 6.46 & 0.54 & 0.88 & 35.60 & 1.12 & 3.13 & 3.56 & 0.61 & 1.06 & \text{1.02} \\
        & \rev{(7.68)} & \rev{(0.74)} & \rev{(1.15)} & \rev{(30.19)} & \rev{(0.88)} & \rev{(3.90)} & \rev{(3.40)} & \rev{(0.67)} & \rev{(1.50)} & \rev{(1.20)} \\ \hline

        \multirow{2}{*}{PE+PP+PVC} & 7.11 & 0.48 & 0.86 & 8.40 & 0.63 & 1.09 & 3.90 & 0.63 & 1.09 & 1.07 \\ 
        & \rev{(8.54)} & \rev{(0.69)} & \rev{(1.18)} & \rev{(7.30)} & \rev{(0.58)} & \rev{(1.03)} & \rev{(3.63)} & \rev{(0.75)} & \rev{(1.28)} & \rev{(1.12)} \\

    \hline
    \end{tabular}}

    \label{tab:mixed_polymer}
\end{table}

\begin{figure}[h]
 \centering
\subfloat[PE.]{\includegraphics[trim = 30mm 90mm 40mm 90mm, clip=true,width=0.33\textwidth]{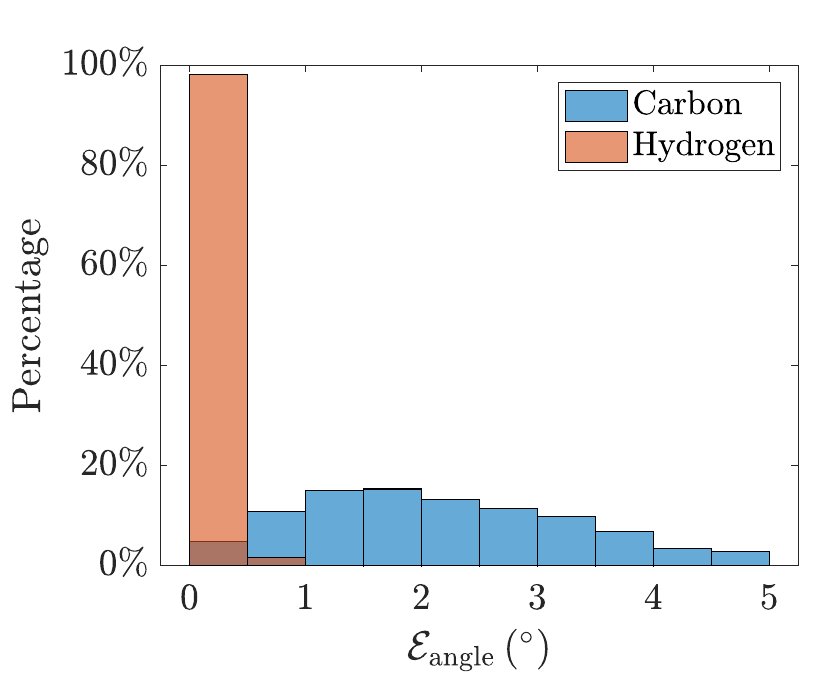}}
\subfloat[PP.]{\includegraphics[trim = 30mm 90mm 40mm 90mm, clip=true,width=0.33\textwidth]{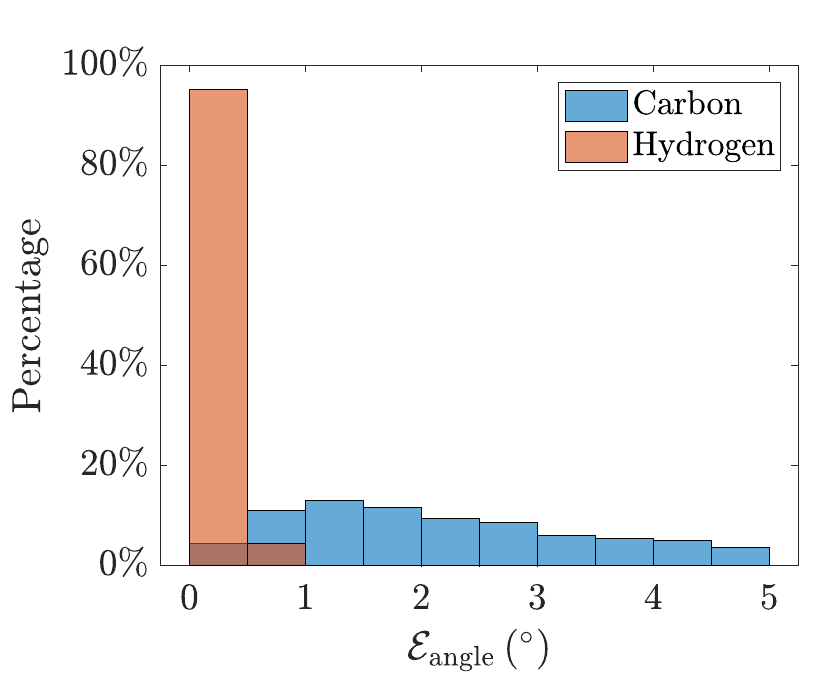}}
\subfloat[PVC.]{\includegraphics[trim = 30mm 90mm 40mm 90mm, clip=true,width=0.33\textwidth]{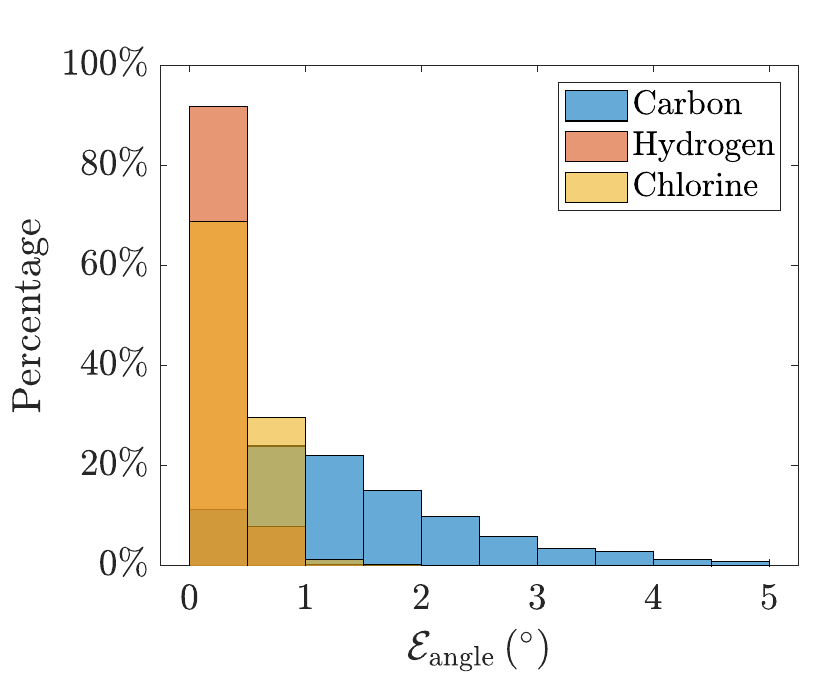}}
    \caption{Histogram of angular error ($\mathcal{E}_\text{angle}\,(^\circ)$) for the model trained on each single-polymer dataset and tested on the corresponding test dataset.} \label{fig:angle}
\end{figure}

For single-polymer datasets, the model achieves low errors on the corresponding test datasets, with higher relative errors observed for carbon atoms. This is primarily due to their smaller force magnitudes and lower representation in the polymer structure compared to other atoms. Among the three polymers, the PVC case exhibits lower errors on carbon atoms, attributed to its distinct atomic composition and slightly higher force magnitudes of carbon atoms. However, these errors remain negligible for molecular simulations, given the larger mass of carbon atoms. Notably, compared to a PW model, the MBD model not only alters the magnitude of vdW dispersion forces but also significantly twists their directions due to many-body correlations \cite{MBD_Polymer_SOSA}. This directional alteration is crucial for accurately capturing many-body effects in the dynamical behavior of polymer melts, emphasizing the importance of force direction predictions alongside magnitudes. To assess this, we further examine the atomic force direction (angular) error $\mathcal{E}_\text{angle}\,(^\circ)$, as defined in Eq.~\eqref{eq:e_angle}, for models trained on single-polymer datasets, corresponding to the bold results in Tab.~\ref{tab:mixed_polymer}. As shown in Fig.~\ref{fig:angle}, force predictions for hydrogen atoms exhibit the highest angular accuracy across all three polymers, while chlorine atoms in PVC also maintain an angular error below $1^\circ$, in agreement with the observed $\mathcal{E}_\text{MARE}$. For carbon atoms, due to the aforementioned limitations, the angular error is generally higher, predominantly ranging between $1^\circ$ and $5^\circ$. Moreover, the model trained on PVC performs best on carbon atoms, aligning with the trend observed in the magnitude-based evaluation. Given the strong correlation and consistency between $\mathcal{E}_\text{MARE}$ and $\mathcal{E}_\text{angle}$, we do not present angular error analysis for the remaining cases.

\begin{figure}[h]
\centering
\includegraphics[trim = 0mm 90mm 0mm 90mm, clip=true,width=0.6\textwidth]{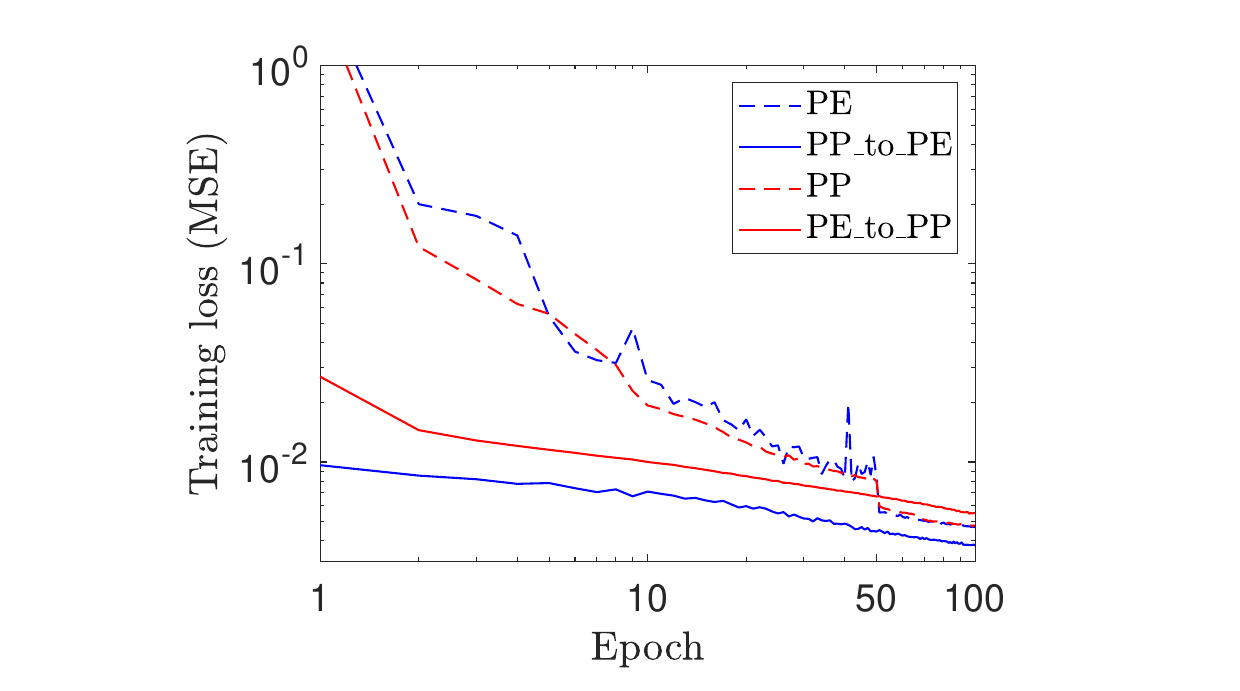}
    \caption{Training convergence of the transfer learning experiment. Solid lines represent transfer learning (learning rate $10^{-4}$) between PE and PP: the PP-trained model fine-tuned on PE (blue) and the PE-trained model fine-tuned on PP (red). Dashed lines in corresponding colors indicate standalone training, which begins with a learning rate of $10^{-3}$ for the first 50 epochs and changes to $10^{-4}$ for the last 50 epochs.}
     \label{fig:transfer_learning}
\end{figure}

Regarding the generalization and transferability of our model, we first evaluate its performance on unseen polymers. In this scenario, the model trained on PP demonstrates superior generalization to PE, likely due to PP’s more complex structure, which enriches its feature embeddings while retaining similarities to PE. This observation is further supported by a transfer learning experiment, as shown in Fig.~\ref{fig:transfer_learning}, where the PP-trained model quickly adapts to the PE dataset, reaching a lower loss at 100 epochs compared to standalone training. In contrast, the PE-trained model requires more epochs to converge when transferred to PP, indicating a more challenging adaptation process. The generalization capability is further assessed by training on mixed datasets with equal ratios of different polymers, specifically: 36k PE + 36k PP, 36k PE + 36k PVC, and 24k PE + 24.3k PP + 24k PVC. For testing, we use the same test datasets as in the single-polymer cases. Although the monomer batching strategy introduced earlier is not applicable, the results in Tab.~\ref{tab:mixed_polymer} demonstrate the model’s capability to generalize across multiple polymer types. Models trained on mixed datasets achieve competitive test errors across all individual polymers, highlighting their potential for applications involving diverse polymer systems.

\rev{In addition to the mean error values, Tab.~\ref{tab:mixed_polymer} includes their standard deviations in parentheses. In most cases, the standard deviations are comparable to or larger than the mean values due to the skewed and heavy-tailed nature of the error distributions, as confirmed by the histograms presented in Fig.~\ref{fig:angle} in this section and Fig.~\ref{fig:rsscs} in Section~\ref{sec:advanced_MBD}. Such distributions reflect the intrinsic complexity of the MBD force landscape and configurational diversity in polymer melt systems. For clarity, standard deviations are omitted in subsequent tables, as they focus on model comparisons and the overall statistical trends remain consistent.}

\subsection{Analysis of key components in trimmed SchNet}
\label{sec:Model_analysis}

\subsubsection{Effects of extra connections}
\label{sec:extra_connnections}
\rev{Regarding the extra connections, we use $p=2$ neareast neighbors and $N_\text{extra}=50$ for all cases. This introduces only $10\%$ more connections to be encoded, requiring no additional parameters and having a negligible impact on computational cost. As shown in Tab.~\ref{tab:extra_connection}, $p=2$ provides the most consistent accuracy across all polymer types, offering notable improvements over no extra connection ($p=0$) or only $p=1$ nearest neighbor. While $p=3$ may yield slight gains for certain cases, the changes are marginal under the same training strategy. The enhancement from these extra connections aligns with the many-body nature of the fitting target: by incorporating extra connections closer to the center, even if not directly linked to it, the model can more effectively capture finer details of the local atomic environment through enriched feature embeddings of nearby atoms. This is analogous to the coupling of dipole–dipole interactions in MBD via diagonalization. In our neural network model, instead, interactions are ``coupled'' through the non-linearity of the proposed shifted softplus activation function, as shown in Eq.~\eqref{eq:ssp}.}

\begin{table}[h]
    \caption{Effects of extra connections on model performance ($\mathcal{E}_\text{MARE}\,(\%)$) for different polymers. \rev{Results are shown for varying $p$ values (number of nearest neighbors) with $N_\text{extra}$ fixed at 50.}}
    \centering
    \resizebox{\textwidth}{!}{
    \begin{tabular}{|c|c|c|c|c|c|c|c|c|c|c|c|c|c|c|c|c|}
    \hline
        \rev{p} & \multicolumn{4}{c|}{0} & \multicolumn{4}{c|}{1}& \multicolumn{4}{c|}{2}& \multicolumn{4}{c|}{3}  \\ \hline
        Test & C & H & Cl & all & C & H & Cl & all & C & H & Cl & all & C & H & Cl & all \\ \hline
        PE & 8.45 & 0.65 & - & 1.10 & \rev{6.67} & \rev{0.46} & - & \rev{0.81} & {5.59} & \textbf{0.41} & - & \textbf{0.71} & \rev{\textbf{5.43}} & \rev{0.46} & - & \rev{0.74}  \\ \hline
        PP & 7.48 & 0.64 & - & 1.04 & \rev{10.22} & \rev{0.54} & - & \rev{1.10} & {6.38} & \textbf{0.49} & - & \textbf{0.83} & \rev{\textbf{6.20}} & \rev{0.56} & - & \rev{0.89}  \\ \hline
        PVC & 4.39 & 0.80 & 1.25 & 1.28 & \rev{3.82} & \rev{\textbf{0.55}} & \rev{\textbf{0.93}} & \rev{0.97} & {3.42} & {0.60} & {0.98} & {1.04} & \rev{\textbf{3.34}} & \rev{0.57} & \rev{0.99} & \rev{\textbf{0.95}} \\ 
    \hline
    \end{tabular}}
    \label{tab:extra_connection}
\end{table}

\subsubsection{Performance of trainable rbfs}
\label{sec:rbf_analysis}
\begin{figure}[h]
 \centering
\subfloat[$N_\text{rbf}=10$.]{\includegraphics[trim = 30mm 90mm 40mm 90mm, clip=true,width=0.33\textwidth]{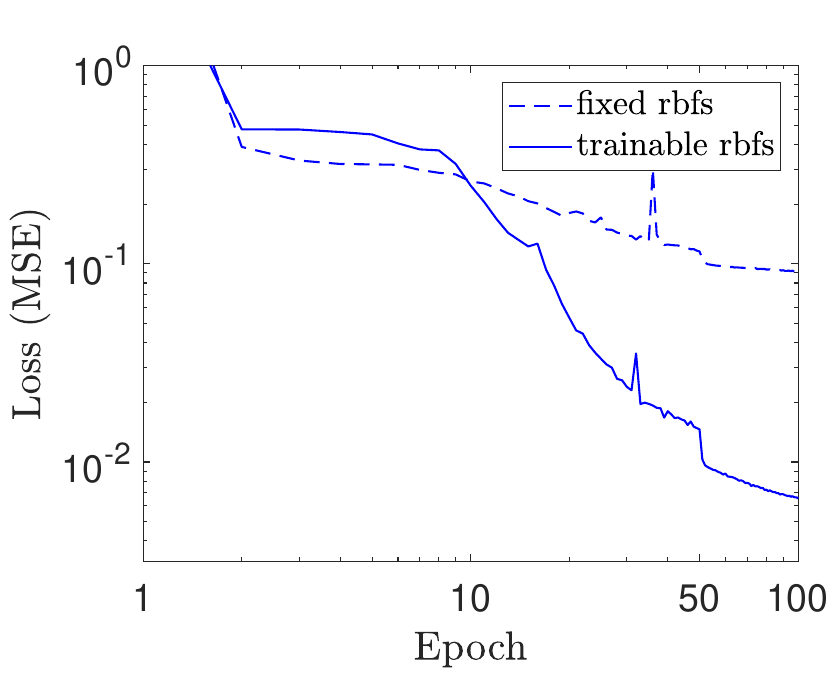}\label{fig:train_PE_Nrbf10}}
\subfloat[$N_\text{rbf}=20$.]{\includegraphics[trim = 30mm 90mm 40mm 90mm, clip=true,width=0.33\textwidth]{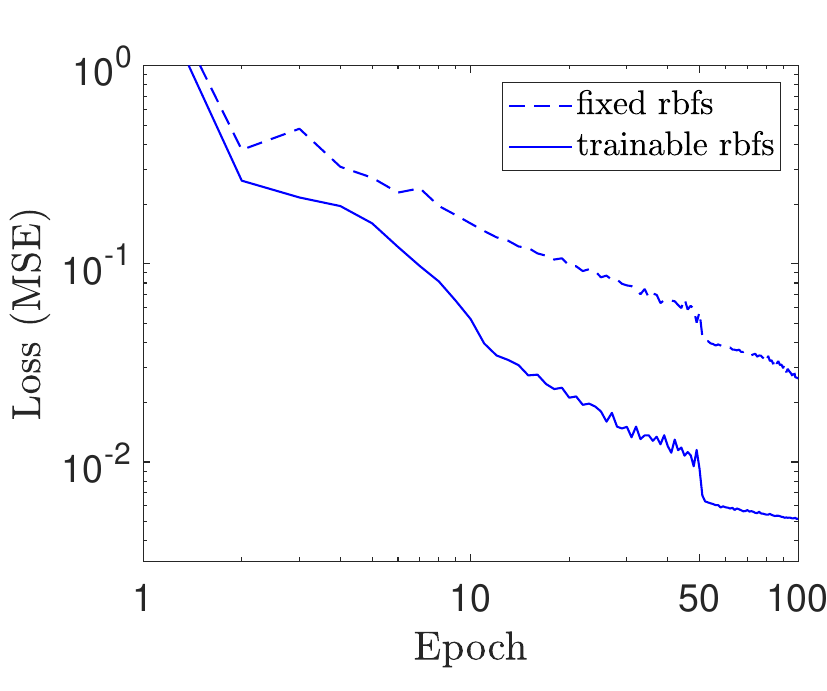}\label{fig:train_PE_Nrbf20}}
\subfloat[$N_\text{rbf}=100$.]{\includegraphics[trim = 30mm 90mm 40mm 90mm, clip=true,width=0.33\textwidth]{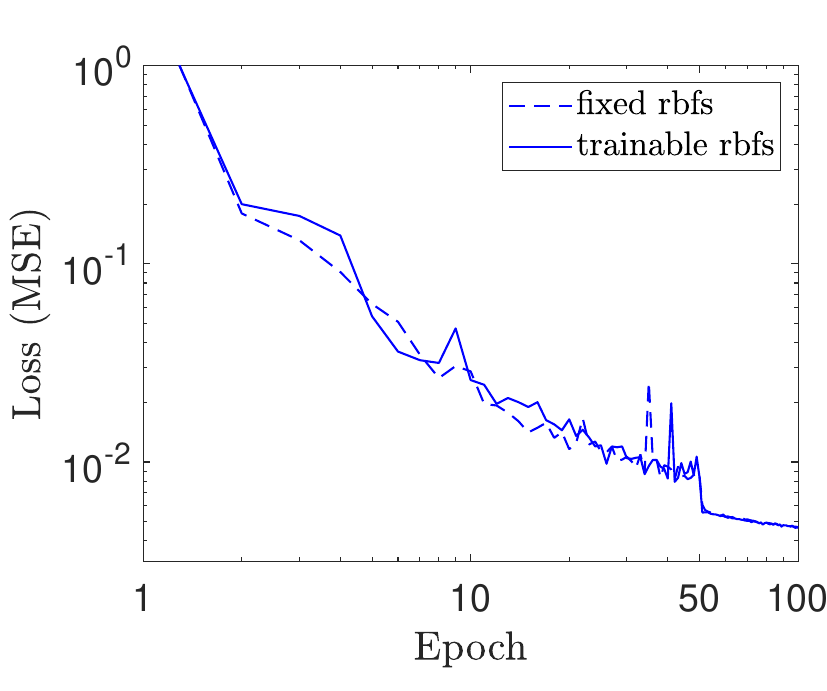}\label{fig:train_PE_Nrbf100}}
    \caption{\rev{Training convergence of the model on PE dataset using different rbf encoding strategies. Three $N_\text{rbf}$ values are considered. Solid curves represent trainable rbfs and dashed curves represent fixed rbfs.}} \label{fig:train_Nrbf}
\end{figure}

For the rbf encoding, the original SchNet uses a single constant value for all $\gamma_k$ and places $\mu_k$ at uniform intervals of $0.1\,\text{\AA}$ across the range of interest. This approach is reasonable for general molecular datasets, where structural configurations or densities vary significantly. However, for a polymer cluster, as shown in Fig.~\ref{fig:PE_cluster}, evenly and finely distributed encoding bases are excessive, given the more consistent spherical geometry, the center-focused target, and the decaying behavior of vdW forces. In this work, we set $N_\text{rbf} = 100$, sufficient to cover all cases, and make both $\gamma_k$ and $\mu_k$ trainable to optimize the encoding strategy. \rev{While trainable rbfs may not significantly improve model accuracy for large $N_\text{rbf}$, since finer fixed bases can adequately encode structure if computational resources are not constrained, they can accelerate training convergence for smaller $N_\text{rbf}$. In particular, for $N_\text{rbf}=10$ and $20$, the trainable rbfs exhibit a much faster loss decay and achieve lower final training loss compared to fixed rbs, as shown in Fig.~\ref{fig:train_Nrbf}. Here, the fixed rbf uses $\gamma_k = 10$ and $\mu_k$ uniformly spaced between $0$ and $15\,\text{\AA}$, which also serves as the initialization for the trainable case. This ability to maintain reasonable accuracy with very few rbfs enables the development of extremely lightweight models, as demonstrated in Tab.~\ref{tab:N_rbf}, whereas models with fixed rbfs under the same settings experience a substantial loss in predictive power.}

\begin{table}[h]
    \caption{Effects of $N_\text{rbf}$ and trainable rbfs on model performance ($\mathcal{E}_\text{MARE}\,(\%)$) for PE.}
    \centering
    \begin{tabular}{|c|c|c|c|c|c|c|c|c|c|}
    \hline
        $N_\text{rbf}$ & \multicolumn{3}{c|}{10} & \multicolumn{3}{c|}{20} & \multicolumn{3}{c|}{100}  \\ \hline
        Test & C & H & all & C & H & all & C & H & all \\ \hline
        Fixed  & 14.92 & 1.98  & 2.71 & 10.15 & 0.89  & 1.40 & \textbf{5.61} & \textbf{0.45}  & \textbf{0.75}\\ \hline
        Trainable & 6.19 & 0.51  & 0.83 & 6.00 & 0.45  & 0.76 &  \textbf{5.59} & \textbf{0.41}  & \textbf{0.71}\\ \hline
    \end{tabular}
    \label{tab:N_rbf}
\end{table}

\begin{figure}[h]
\centering
\includegraphics[trim = 0mm 100mm 0mm 100mm, clip=true,width=1\textwidth]{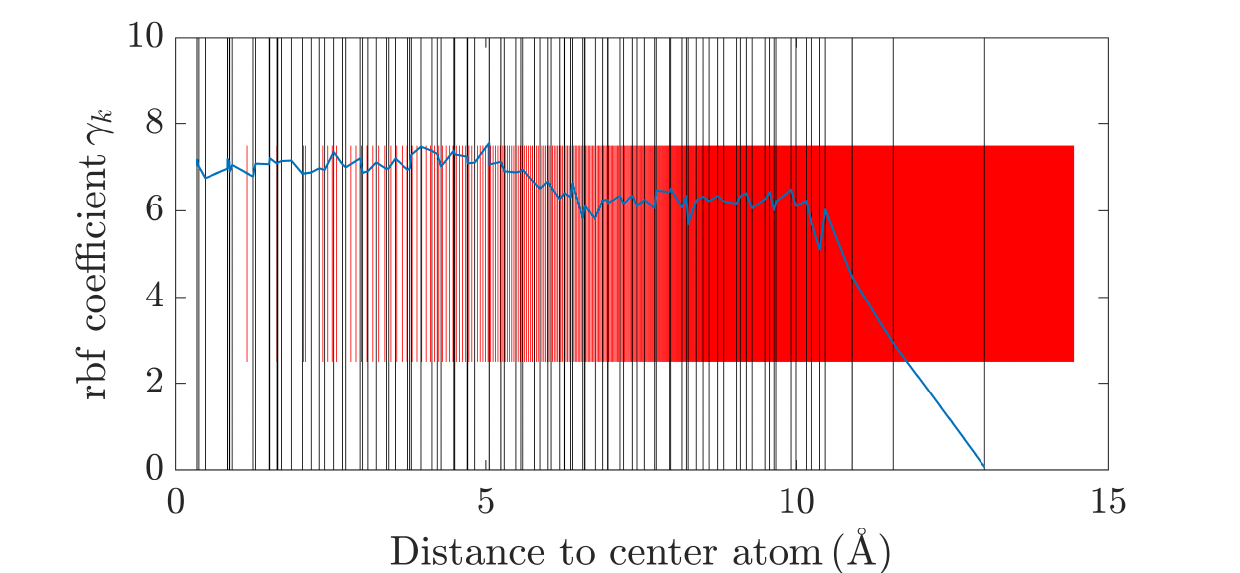}
    \caption{\rev{Visualization of the trained rbfs for PE with $N_\text{rbf}=100$. The x-axis shows distance relative to the center atom, and the y-axis shows rbf coefficient $\gamma_k$. Black vertical lines indicate the learned positions of the rbf centers $\mu_k$, red lines mark the averaged interatomic distances for each neighbor relative to the center atom, and the blue curve illustrate the corresponding $\gamma_k$ values, controlling the width of each basis function.}}
     \label{fig:rbf}
\end{figure}

Moreover, with sufficient trainable rbfs, SchNet encodes atomic structures in a more physically meaningful way, enhancing the model's robustness. As shown in Fig.~\ref{fig:rbf}, rbf centers cluster naturally in the short- and mid-range distances \rev{(see black vertical lines)}, with only a few extending to the long range, where they are weighted by decaying $\gamma_k$ values over distance \rev{(see the blue line)}. In the proposed Gaussian-type rbf, the coefficient $\gamma$ controls the width of the function, and a small $\gamma$ allows the rbf to attend a wider range of neighbors. The observed distribution pattern reveals that the model autonomously identifies regions with the most predictive information and averages long-range contributions without requiring fine details. Notably, at short distances, where nearest-neighbor interactions dominate, an intriguing pattern can be observed: \rev{rbf centers cluster between interatomic distances (red vertical lines)} to better capture local structural features while avoiding zero gradients, as governed by Eq.~\eqref{eq:rbf}.

\subsubsection{Training with unit-specific batching}
\label{sec:batching}
In the extreme case where $N_\text{extra}$ includes all atoms in the system, the setup effectively merges the $p+1$ trimmed connection graphs, allowing direct force predictions for all $p+1$ atoms. With a limited number of $p$ neighbors, the concerning atoms share the same local atomic environment. As a result, using a model with a single common geometry input and a multi-force output, such as predicting the forces of a methylene group collectively, could potentially improve accuracy. This is attributed to the more complete connection graph and a more explicit ``coupling'' achieved during the output phase through gradient descent optimization of the combined force predictions. However, this approach has practical limitations. It is incompatible with the neighbor-search routines in commonly used MD packages, such as JAX\_MD \cite{jaxmd2020_proceeding, jaxmd2021}, where atomic geometries are not necessarily ordered in a desired manner that can align with the fixed multi-force output. Additionally, edge atoms in polymer chains require special treatment, since they do not conform to the fixed-size output structure, making it more complicated for implementation.

\begin{table}[h]
    \caption{Effects of batching strategies on model performance ($\mathcal{E}_\text{MARE}\,(\%)$) for different polymers.}
    \centering
    \begin{tabular}{|c|c|c|c|c|c|c|c|c|}
    \hline
        Batching & \multicolumn{4}{c|}{normal} & \multicolumn{4}{c|}{unit-specific}  \\ \hline
        Test & C & H & Cl & all & C & H & Cl & all \\ \hline
        PE & \textbf{5.53} & 0.42 & - & 0.71 & {5.59} & \textbf{0.41} & - & \textbf{0.71}\\ \hline
        PP & 7.2 & 0.52 & - & 0.91 & \textbf{6.38} & \textbf{0.49} & - & \textbf{0.83}\\ \hline
        PVC & 3.73 & 0.66 & 1.04 & 1.07 & \textbf{3.42} & \textbf{0.60} & \textbf{0.98} & \textbf{1.04}\\ 
    \hline
    \end{tabular}
    \label{tab:unit_batching}
\end{table}

To simulate the multi-force output scenario while keeping a single-output model, we introduce a unit-specific batching strategy during training. This approach groups atoms from the same smallest repeating unit into a single batch. For example, in the training for PE, each batch of size 36 contains 12 complete ethylene groups ($\text{CH}_2$), while for PP, the same batch size includes 4 complete propylene monomers ($\text{C}_3\text{H}_6$). By grouping atoms this way, the optimizer can more effectively ``couple'' them, as the model updates them collectively within the same optimization step. Although this technique does not upgrade the model itself to predict multiple outputs, it smoothens the optimization landscape by leveraging a physically meaningful averaging across each batch. As shown in Tab.~\ref{tab:unit_batching}, the largest error reduction occurs for PP (9-atom unit), while the effect is marginal for PE (3-atom unit). This further demonstrates the many-body nature of the force, where coupling larger atomic groups leads to a more accurate description.

\subsection{Performance for the advanced MBD model}
\label{sec:advanced_MBD}
The MBD method introduced in Section~\ref{sec:mbd} is commonly referred to as the ``plain'' MBD. Building on the same methodology, several MBD variants have been developed, including MBD@rsSCS \cite{ambrosetti2014long}, \rev{universal MBD (uMBD)} \cite{uMBD}, and \rev{nonlocal MBD (MBD-NL)} \cite{NLMBD}. Among these, the MBD@rsSCS version (range-separation self-consistent screening) aligns with the scope of this work that is mainly focused on organic systems. MBD@rsSCS is an advanced formulation designed to seamlessly integrate with \rev{Density Functional Theory (DFT)-based} models, providing a more accurate treatment of polarizability screening across different ranges. This is achieved by applying the range separation technique to the dipole-dipole interaction tensor $\boldsymbol{T}$, allowing a proper derivation of short-range screening and long-range correlation.

\begin{figure}[h]
 \centering
\subfloat[MBD (plain).]{\includegraphics[trim = 30mm 90mm 30mm 90mm, clip=true,width=0.49\textwidth]{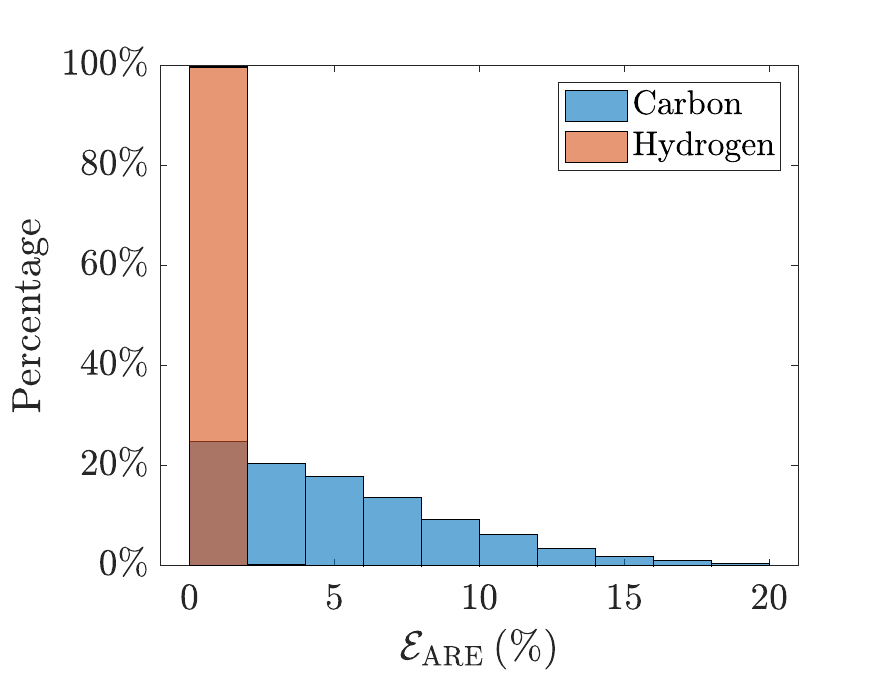}}
\subfloat[MBD@rsSCS.]{\includegraphics[trim = 30mm 90mm 30mm 90mm, clip=true,width=0.49\textwidth]{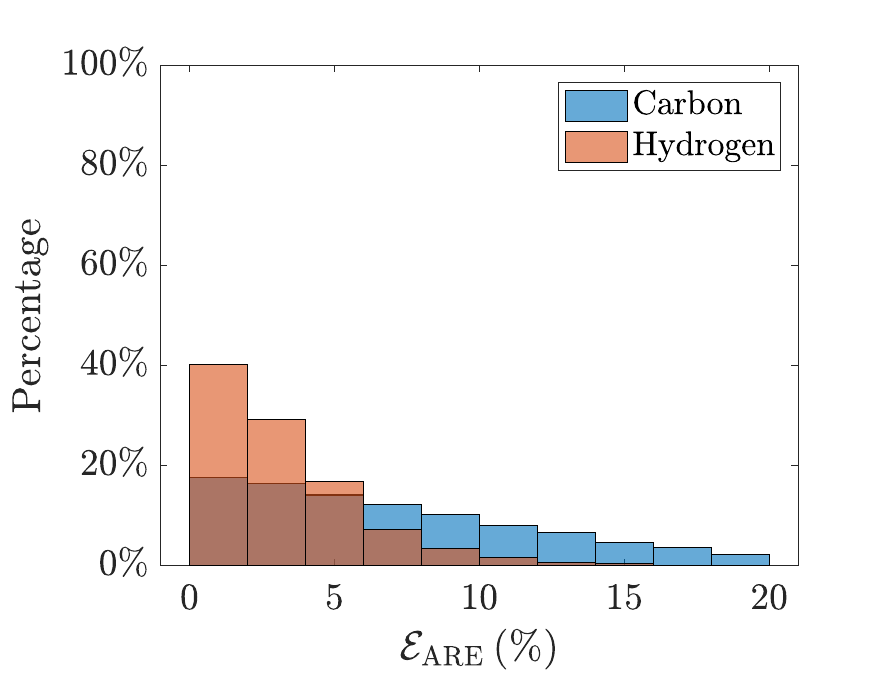}}
    \caption{Histogram of $\mathcal{E}_\text{ARE}\,(\%)$ of the model for two MBD variants in PE.} \label{fig:rsscs}
\end{figure}

For the surrogate modeling, MBD@rsSCS represents a more complex and challenging target compared to the plain version, with an expected loss in predictive accuracy. To explore this, we collected an MBD@rsSCS dataset for PE, revealing notable significant changes in force magnitudes (generally smaller) and distributions, with the forces on carbon and hydrogen atoms becoming closer in magnitude. As shown in Tab.~\ref{tab:rsscs}, the SchNet model trained on MBD@rsSCS forces, with the same architecture and training strategy (scaling forces by $10^4$ instead), exhibits a decreased ability to predict these forces, particularly for hydrogen atoms. The error for hydrogen increases from 0.41\% to 3.25\%, reflecting their reduced force values relative to carbon. The near-2:1 error ratio aligns with the atomic composition in the dataset. This disparity is further illustrated in the error histogram in Fig.~\ref{fig:rsscs}, where $\mathcal{E}_\text{ARE}\,(\%)$ is used as the error metric  (see Eq.~\eqref{eq:e_ARE}). The results show that hydrogen errors for MBD@rsSCS show significantly lower concentration near zero compared to the plain MBD results, while the distribution of carbon remains similar.

\begin{table}[h]
    \caption{Performance ($\mathcal{E}_\text{MARE}\,(\%)$) of the model for two MBD variants in PE.}
    \centering
    \begin{tabular}{|c|c|c|c|c|c|c|c|}
    \hline
        Model & \multicolumn{3}{c|}{MBD (plain)} & \multicolumn{3}{c|}{MBD@rsSCS}  \\ \hline
        Test & C & H & all & C & H & all \\ \hline
        PE & 5.59 & 0.41 & 0.71 & 7.61 & 3.25 & 4.49 \\
    \hline
    \end{tabular}

    \label{tab:rsscs}
\end{table}

\rev{The reduced performance of the model on MBD@rsSCS can be attributed to the increased intrinsic complexity of this variant. Compared to the plain MBD, MBD@rsSCS introduces short-range self-consistent screening and modifies the dipole interaction tensor at long range \cite{ambrosetti2014long}, both of which significantly alter the overall force landscape and force magnitudes. These effects likely make the learning task more challenging. To achieve a similar level of accuracy as in the plain MBD case, we would need to increase the size of the model or apply dedicated hyperparameter tuning to better capture the more complex force distribution.} However, this does not indicate a limitation but rather demonstrates the expected trade-off when addressing a more complex problem. Essentially, this test demonstrates the robustness of our trimmed SchNet in handling altered force distributions, highlighting its adaptability. Furthermore, it motivates the exploration of new techniques to accommodate varying levels of target complexity in the future. Within the scope of this work, where we aim to develop a surrogate model for MBD that integrates with general force field models for molecular simulations, the plain MBD fits better than other MBD variants. Hence, the current trimmed SchNet adequately meets our requirements for this application.

\subsection{Preliminary study of temperature effects}
\label{sec:temperature}
Despite the strong predictive capability of the proposed trimmed SchNet model for our polymer melt dataset, we acknowledge that the obtained melt structures may not fully represent realistic crystallization behaviors. Ideally, at the same equilibration temperature of 300\,K, the three types of polymer should exhibit varying degrees of crystallinity due to their different structural and thermal properties. However, this distinction is likely underrepresented in the dataset due to the limited equilibration time steps permitted by the online CHARMM-GUI. Nevertheless, we are still able to conduct a preliminary study on the effects of temperature.

\begin{table}[h]
    \caption{Performance ($\mathcal{E}_\text{MARE}\,(\%)$) of the model trained on mixed-temperature dataset of PE.}
    \centering
    \begin{tabular}{|c|c|c|c|c|c|c|}
    \hline
        \multirow{2}{*}{\diagbox{Train}{Test}} & \multicolumn{3}{c|}{mixed-size} & \multicolumn{3}{c|}{mixed-temperature}  \\ \cline{2-7}
         & C & H & all & C & H & all  \\ \hline
        mixed-size & \textbf{5.59} & \textbf{0.41} & \textbf{0.71} & \textbf{5.50} & 0.43 & 0.71\\ \hline
        mixed-temperature & 5.90 & 0.45 & 0.76 & {5.52} & \textbf{0.42} & \textbf{0.71} \\ 
    \hline
    \end{tabular}

    \label{tab:mixed_temperature}
\end{table}

Here, we focus on one configuration of PE and construct an additional mixed-temperature dataset. We equilibrate an $8\times300$ PE melt at five different temperatures, ranging from 100\,K to 300\,K, which leads to mild volume changes, with the unit cell length spanning from 55.14\,$\text{\AA}$ to 56.41\,$\text{\AA}$. From this dataset, we again collect 60k data points for training and reserve an additional 6k for testing. The model is trained on this mixed-temperature dataset using the same training strategy as in previous cases. As shown in Tab.~\ref{tab:mixed_temperature}, we cross-compare this newly trained model with the model trained on the original dataset introduced in Section~\ref{sec:data}, which is equilibrated at 300\,K but includes different chain lengths (referred to as the mixed-size dataset). The two models exhibit similar performance when tested on both datasets, despite the fact that $1/5$ of their data points are identical.

These results demonstrate that the proposed model generalizes well across polymer chain lengths and minor density variations. The consistent accuracy, even across a wide temperature range, further suggests that the representation of crystallinity in the mixed-temperature dataset is limited, leading to predominantly amorphous structures. This, in turn, may enhance the regularity of the cutoff cluster, thereby simplifying the surrogate modeling task. \rev{We acknowledge that a more rigorous validation would require testing on systems with pronounced crystallinity, which could be achieved through longer equilibration or dedicated crystalline sampling. Such systems will be included in future work to evaluate the model’s robustness under reduced structural regularity and the presence of ordered domains.} 

\subsection{Analysis of the Hessian}
\label{sec:hessian}
\begin{figure}[h]
 \centering
\subfloat[PE.]{\includegraphics[trim = 30mm 90mm 40mm 90mm, clip=true,width=0.33\textwidth]{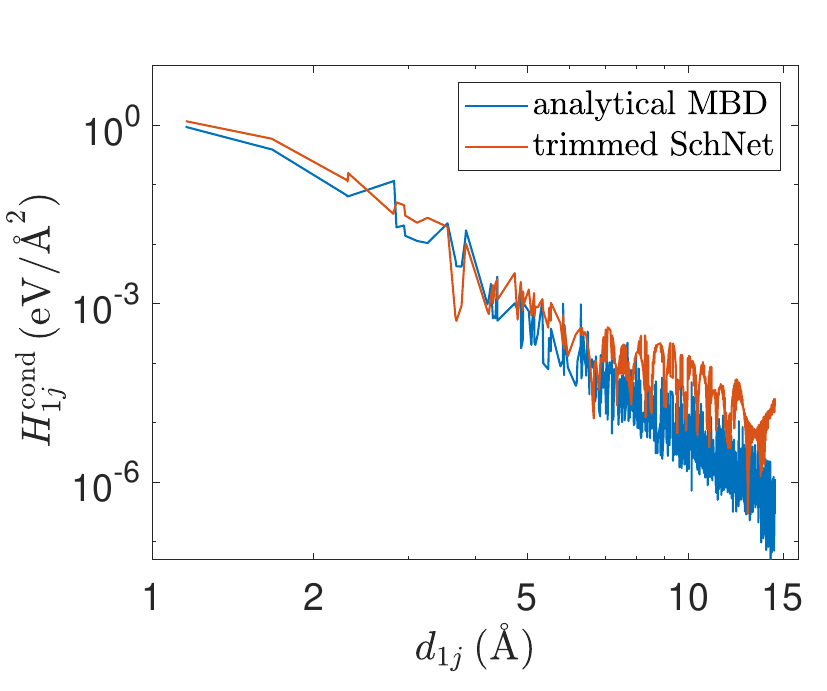}\label{fig:hessian_PE}}
\subfloat[PP.]{\includegraphics[trim = 30mm 90mm 40mm 90mm, clip=true,width=0.33\textwidth]{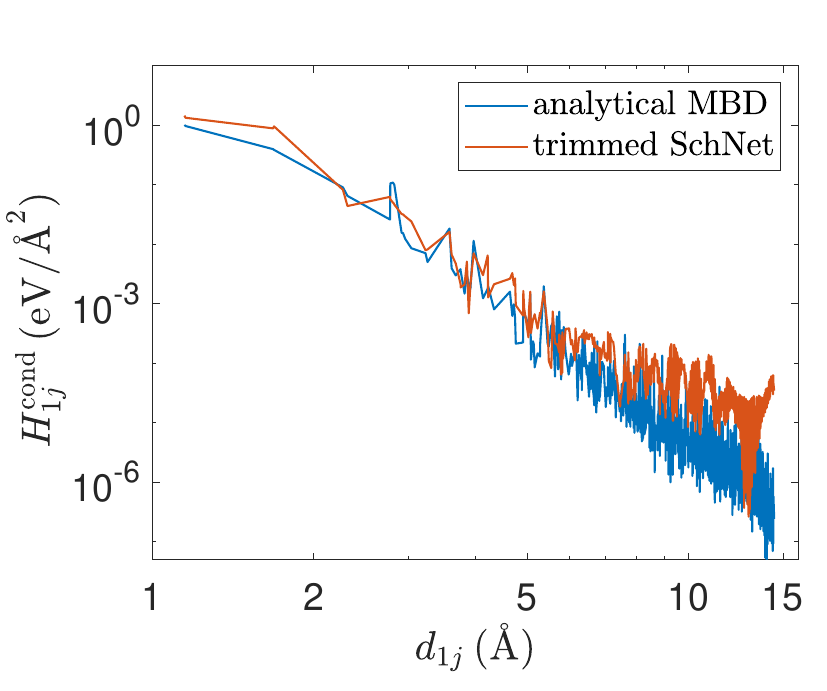}\label{fig:hessian_PP}}
\subfloat[PVC.]{\includegraphics[trim = 30mm 90mm 40mm 90mm, clip=true,width=0.33\textwidth]{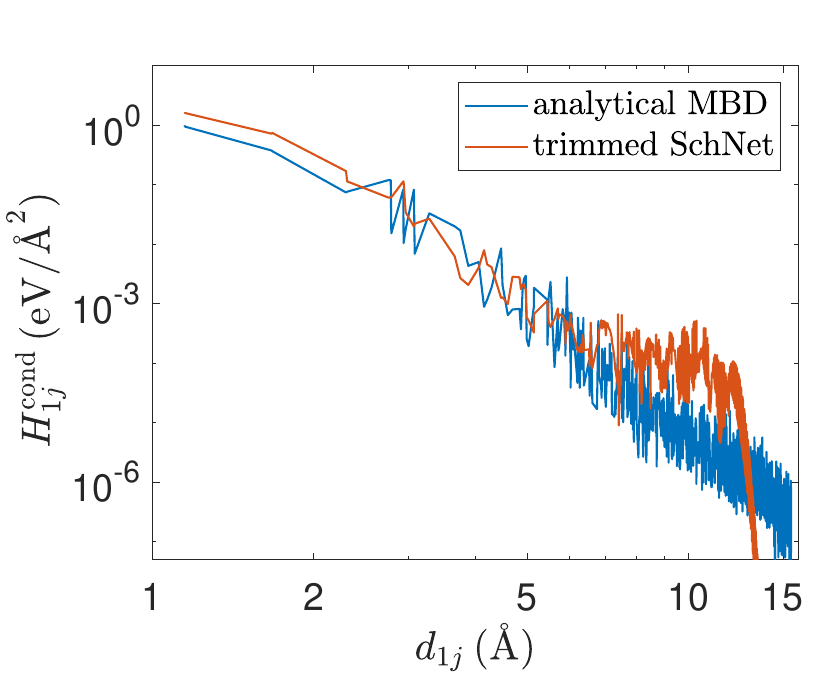}\label{fig:hessian_PVC}}
    \caption{Condensed Hessian profiles for a random carbon atom in different polymers. In the figures, we order ${H}_{1j}^\text{cond}$ in terms of the corresponding interatomic distance to the center atom $d_{1j}$ for clarity.} \label{fig:hessian}
\end{figure}

Differing from generic deep neural network models, such as fully connected neural networks, SchNet incorporates a more physically meaningful architecture by applying convolutional operations to pairwise atomic connections. This is analogous to the analytical MBD force formulation, where the dipole matrix $\boldsymbol{C}^\text{MBD}$ undergoes differentiation and scaling through eigen-decomposition products. Similarly, SchNet effectively scales the convolved atomic connections through activations and computes forces via a gradient operation. Thanks to the infinite continuity of softplus activation functions and TensorFlow’s AD function, it is possible to extract the second derivative of a trained SchNet model, corresponding to the Hessian $\boldsymbol{H}_{1j}$. This tensor serves as a valuable tool for analyzing atomic interactions, as it quantifies the force response on a center atom due to infinitesimal displacements of neighboring atoms within the cutoff range \cite{Colossal_hessian}. In Fig.~\ref{fig:hessian}, we evaluate the Frobenius norm of $\boldsymbol{H}_{1j}$, referred to as the ``condensed Hessian'' ${H}_{1j}^\text{cond}=\|\boldsymbol{H}_{1j}\|_\text{F}$, and compare the condensed Hessian profiles obtained from the analytical MBD model and the trained trimmed SchNet model for a randomly chosen carbon atom in different polymer melts. \rev{This condensed form serves as a scalar representation for the effective force constants associated with MBD interactions, offering a physically intuitive and easily visualizable measure of how MBD stiffness decays with interatomic distance.} For all cases, wavy behaviors are observed at long range, attributed to many-body correlation effects and varying atom types over distance. In the analytical Hessian profiles, we find a polynomial decay power below 7, which is lower than the decay power of 8 for the PW case. This further underscores the non-trivial modifications introduced by MBD to the power laws that govern the vdW dispersion interactions \cite{wavelike_science_2016}, while also justifying the need for an extended cutoff distance for MBD force calculations.

Acknowledging the physical significance of the Hessian, our model effectively reproduces the decaying behavior observed in the analytical reference and loosely captures the intricate short-range details, further demonstrating the network's robustness. Notably, the Hessian profile for the PVC system reveals a rapid decay near the cutoff boundary, suggesting that the current $N_\text{cut}$ is excessive for this polymer. This observation is reasonable given that the longer $\text{C–Cl}$ bonds result in a larger radius for the cutoff cluster, thus reducing the number of atoms required for calculating MBD forces. This analysis provides valuable insights for optimizing the cutoff strategy, allowing us to tailor the cutoff range more precisely for different polymer systems.

\subsection{Towards implementation for MD simulations}
\label{sec:MD}
We have open-sourced the polymer melt dataset \cite{polymer_dataset} and the TensorFlow code of the trimmed SchNet architecture \cite{trimmedSchNetRepo} to facilitate future research on surrogate modeling of MBD interactions. One of the most direct use of our surrogate model would be its integration in MD simulations.  However, such an integration can be challenging, and in this section we discuss possible approaches to do it efficiently. 

Technically, as the MD simulation framework we propose using JAX\_MD, with the trimmed SchNet model introduced as a custom potential implemented with Flax\_NNX \cite{flax2020github}. Both libraries provide JAX-based APIs \cite{jax2018github}, enabling a smooth and straightforward integration of the neural network into a standard MD simulation. As demonstrated in the demo provided in our repository, the user simply needs to define a force function that calls the SchNet model to calculate the MBD forces for a given atomic system $\boldsymbol{R}$. Due to VRAM limitations, inference is performed iteratively in a for-loop, processing subsets of the full system. 

To estimate the computational performance of the model, we have carried out an NVT simulation of a 9k-atom system on a 32GB Nvidia V100 GPU (with an inference batch size of 3000). The model predictions achieve a speed of 0.02\,ms/atom/step, leveraging JAX's high-performance computing capabilities. This is several order of magnitude faster to a single analytical MBD calculation for one atom in a polymer melt, which requires approximately 1\,s.

\begin{figure}[h]
\centering
\includegraphics[trim = 0mm 85mm 0mm 90mm, clip=true,width=0.6\textwidth]{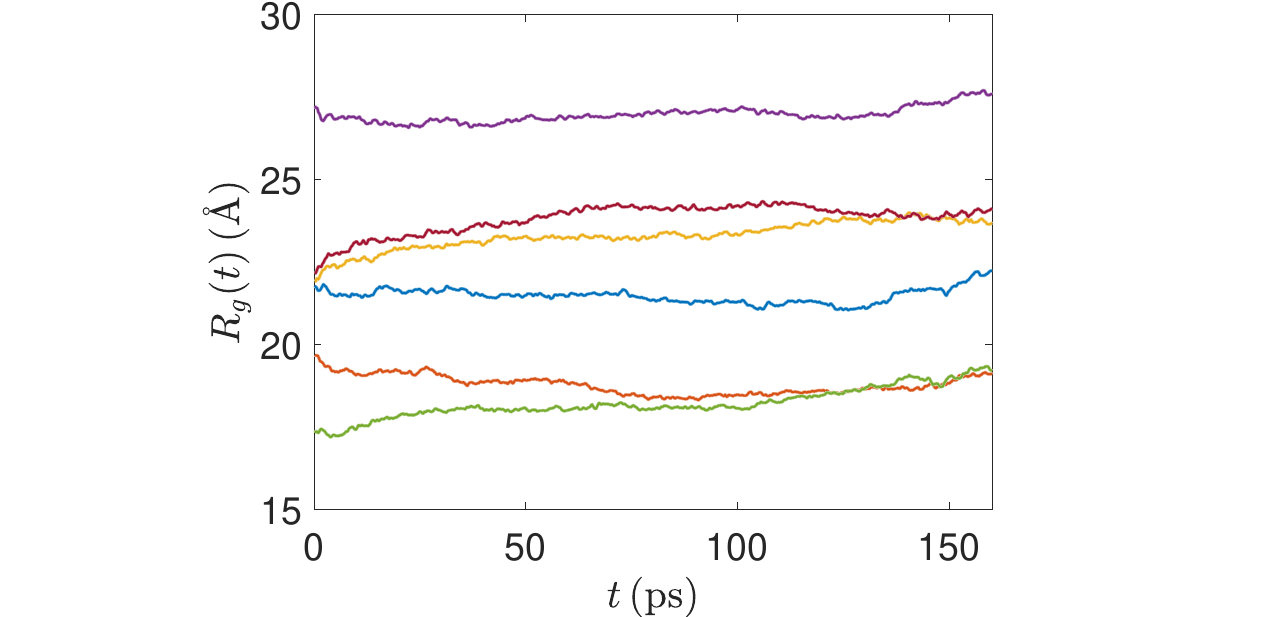}
    \caption{\resub{Time evolution of the radius of gyration $R_g(t)$ for six polymer chains in a PE melt during an NVT MD simulation using the trimmed SchNet MBD surrogate. Each curve corresponds to one polymer chain and represents an average over multiple independent random-seed simulations.}}
     \label{fig:Rg}
\end{figure}

\resub{To demonstrate that the proposed surrogate can be stably integrated into an MD workflow, we performed a preliminary MD test using a practical FF coupling strategy for polymer melts. As discussed in Section~\ref{sec:mbd}, the MBD surrogate model serves as the attractive component of the vdW interaction, and must therefore be combined with additional interaction terms to enable MD simulations. In this context, the surrogate is coupled to TraPPE \cite{Trappe}, a classical FF model for polymers, following the implementation strategy detailed in \cite{theis_SHEN}. The simulation is conducted at 300\,K in the NVT ensemble for a PE system with approximately 9k atoms, consisting six polymer chains. As a simple and widely used structural observable, the time evolution of the radius of gyration $R_g(t)$ is monitored for each chain for around 150~picoseconds. The resulting trajectories shown in Fig.~\ref{fig:Rg} exhibit smooth and bounded behavior over a relatively long simulated time window, indicating that the surrogate model behaves stably when used in a standard MD workflow. The purpose of this test is solely to demonstrate numerical stability and practical integrability of the model, rather than to validate polymer structural or thermodynamic properties.}

The above mentioned practical coupling is in general not sufficient. Deploying the MBD surrogate as part of a complete and quantitatively predictive FF for polymer melts requires the inclusion of bonded interactions and a \emph{consistent} short-range repulsive component for the vdW interaction. In principle, the surrogate can be coupled to an existing or lightly tuned LJ repulsive term within a classical FF to enable rapid, qualitative investigations of MBD effects relative to the PW model. For rigorous, quantitative coupling, however, the repulsive parameters should be refitted using the same physical targets employed in the original force field development. For example, in TraPPE, LJ parameters are calibrated to reproduce vapor--liquid phase equilibria, and replacing the attractive component with the MBD surrogate would naturally be accompanied by re-calibrating the repulsive term to preserve these target properties.


In even longer perspective, a more physically consistent integration can be achieved by coupling the surrogate to a short-range MLFF such as SO3LR or GEMS. These general-purpose MLFFs already model bonded and short-range non-bonded interactions at high fidelity by fitting to DFT-level data, whereas explicit long-range MBD effects are not yet incorporated in these approaches. Moreover, in the context of the range-separation concept as discussed in \cite{DFTB+MBD}, DFT-based models inherently lack a description of long-range correlation and require a vdW dispersion complement. This makes coupling to the MBD surrogate both smooth and physically consistent. Such a combination would yield a complete MLFF with accurate short-range physics and long-range MBD interactions. Developing and validating this coupling, including tests on structural, dynamic, and thermodynamic properties from MD simulations, are in progress but lie beyond the scope of the present manuscript.

\section{Conclusions and future work}
\label{sec:conclusion}
In this paper, we introduced a trimmed SchNet architecture as a surrogate model specifically designed for predicting MBD forces in polymer melts. Despite the widely recognized success of MBD over PW methods in accurately modeling many-body vdW effects, and the rapid growth of MLFFs, \rev{dedicated ML surrogates for MBD remain largely unexplored. Our work addresses this gap by presenting a model tailored for MBD in large-scale molecular simulations.} Polymer melts were chosen as the initial application due to their strong dependence on vdW interactions, and their structural characteristic guided both our modeling strategy and architecture design.

In large-scale MD simulations, vdW forces are computed repeatedly for the center atom within a finite-size atomic cluster across the full polymer melt system, making MBD surrogate predictions follow the same pattern. Due to the near-spherical shape of the cutoff clusters, this surrogate task exhibits surprising regularity, making it suitable for ML modeling. The proposed SchNet-based architecture retains the core geometric encoding and continuous-filter convolution techniques of the original SchNet while introducing key modifications specific to our problem. The trimmed SchNet not only reduces the model size but also simplifies the connection graph by keeping only interactions between the center atom and others, with a few additional connections to the two nearest neighbors of the center. This significantly reduces computational cost while preserving essential many-body correlations. Additionally, the model employs trainable rbf encoding, reducing the number of required basis functions, and incorporates a unit-specific batching strategy that leverages the repetitive nature of polymer monomers to enhance training convergence.

The model was trained on MBD force datasets for PE, PP, and PVC melts, demonstrating high predictive accuracy across all three polymer systems. Furthermore, it generalizes well when trained on a combined dataset of multiple polymers. When applied to advanced MBD variants, such as MBD@rsSCS, it reveals the challenges posed by more complex force distributions. In addition, a preliminary study of temperature effects shows that the model maintains accuracy against mild volume changes in the structure, further demonstrating its robustness. Moreover, the trimmed SchNet model exhibits strong physical interpretability. The analysis of the trained model's Hessian shows that it captures the characteristic decay behavior of MBD interactions, offering insights into cutoff optimization for different polymer systems. Finally, we discussed the practical implementation of the model in MD simulations, supported by a demo using the JAX\_MD library, which showcases its applicability to large-scale MD simulations. To maximize the impact of this work, all the codes and datasets used in this study are available in our repository \cite{trimmedSchNetRepo}.

With the proposed trimmed SchNet model, one can already integrate it into MD simulations of large-scale polymer melts, provided that it is consistently coupled with an appropriate FF model for the remaining interactions in the system. While the present work demonstrates the numerical stability and practical feasibility of such integration, achieving a rigorous and quantitatively predictive coupling requires further effort. Our immediate future work will focus on this direction, with particular interest in potential emergent MBD effects that could lead to significant differences in polymer melt morphology and mechanical properties compared to the results from PW vdW models.

However, we acknowledge certain limitations of the proposed architecture. The success of the model is closely tied to the structural characteristics of the obtained polymer melts, where the resulting regularity of the cutoff clusters ensures a well-defined learning task. The impact of crystallinity on predictive performance remains untested. This also suggests that the current model may not generalize well to molecular systems with highly asymmetric and irregular geometries around each target atom, such as large protein molecules. In future work, we will first evaluate the model’s performance in crystalline polymer structures and assess the influence of cutoff regularity. We also plan to extend our approach to more complex polymers and other condensed-phase systems, such as liquids. Ultimately, we aim to generalize the MBD surrogate to a broader class of molecular systems and integrate it into state-of-the-art MLFF frameworks such as SO3LR \cite{SO3LR} and GEMS \cite{GEMS}.

Finally, this work is part of the wider effort within our research group towards ML surrogate modeling at different scales \cite{SO3LR, GEMS, Deshpande_probab_DNN, MagNet,MagNet_perceiver_Frontier, GP_plus_Autoencoder,PAPAVASILEIOU2023103938,Surrogate_elastoplastic} and towards effective multiscale modeling \cite{Hauseux, DFTB+MBD, multiscale_elastoplastic}.  
We hope our contribution and open-source repository will serve the community of computational engineering for future research on \textit{ab initio} modeling of large-scale systems.

\section*{Acknowledgments}
We are grateful for the support of the Luxembourg National Research Fund (C20/MS/14782078/QuaC). The calculations presented in this paper were carried out using the HPC facilities of the University of Luxembourg. 

\appendix
\section{MBD formalism in detail}
\label{app_sec:mbd}

The many-body dispersion (MBD) method \cite{PhysRevLett.108.236402,doi:10.1063/1.4789814} is based on the Adiabatic Connection Fluctuation Dissipation Theorem (ACFDT) within the Random Phase Approximation (RPA) and models the system by quantum harmonic oscillators (QHOs) coupled through dipole-dipole interactions \cite{doi:10.1063/1.4789814}. Unlike PW methods, the MBD model accounts for all orders of dipole-dipole interactions among fluctuating atoms. The ACFDT-RPA correlation energy for the MBD model is expressed as:
\begin{equation}
E^{\textrm{MBD}} = \dfrac{1}{2\pi} \int _0 ^\infty \Tr \left[ \text{ln}(\boldsymbol{1} - \boldsymbol{AT})\right] \text{d} \omega,
\label{eq:E_mbd_ACFDT-RPA}
\end{equation}
where $\boldsymbol{A}$ is a diagonal $3N \times 3N$ matrix for a system of $N$ atoms. For isotropic QHOs, the matrix elements are defined as ${A}_{lm} = -\delta_{lm} \alpha_l \left( i \omega \right)$, with $\alpha_l \left( i \omega \right)$ representing the $l$th frequency-dependent atomic polarizability. $\boldsymbol{T}$ is so called dipole-dipole interaction tensor:
\begin{equation}
T_{ij}^{ab} = \nabla_{\boldsymbol{r}_i} \otimes \nabla_{\boldsymbol{r}_j} v_{ij}^{gg},
\label{ddinter}
\end{equation}
where $a$ and $b$ denote the Cartesian coordinates, and $v_{ij}^{gg}$ is a modified Coulomb potential \cite{KWIK_erf}, incorporating overlap effects between fluctuating point dipoles:
\begin{equation}
v_{ij}^{gg} = \dfrac{\text{erf}\left(  r_{ij}/ (\beta\cdot \tilde{\sigma}_{ij})  \right) }{r_{ij}}.
\label{ModiCoulPo}
\end{equation}
The parameter $\beta$ in Eq.~\eqref{ModiCoulPo} is an empirical constant and $\tilde{\sigma}_{ij} = \sqrt{\sigma_{i}^2 + \sigma_{j}^2}$ represents an effective width derived from the Gaussian widths $\sigma_{i}$ and $\sigma_{j}$ of atoms $i$ and $j$, respectively. These widths are directly linked to the polarizabilities, $\alpha_{i}$, as per classical electrodynamics \cite{PhysRevB.75.045407} and can be computed from the dipole self-energy:
\begin{equation}
\sigma_{i} = \left(\sqrt{\dfrac{2}{9\pi}} \alpha_{i}\right)^{1/3}. 
\label{eq:Gaussian_width}
\end{equation}

To improve computational efficiency, the dipole-dipole interaction is assumed to be frequency-independent using the effective polarizability $\alpha_{i}^{\text{0,eff}}$ to calculate $\sigma_{i}$ in Eq.~\eqref{eq:Gaussian_width}. For a frequency-independent $\boldsymbol{T}$ tensor, the MBD energy can be obtained directly from the eigenvalues $\lambda_p$ of the $3N \times 3N$ matrix $\boldsymbol{C}_{}^{\text{MBD}}$ \cite{doi:10.1063/1.4789814, ambrosetti2014long}, which describe the coupling between each pair of atoms $i$ and $j$ with a $3\times3$ block: 
\begin{equation}
    \boldsymbol{C}_{ij}^{\text{MBD}} = \omega_i^2 \delta_{ij} + (1-\delta_{ij}) \ \omega_i\omega_j \sqrt{ \alpha_{i}^\text{0,eff} \alpha_{j}^\text{0,eff}}\ \boldsymbol{T}_{ij}.
    \label{eq:Cmat}
\end{equation}
where $\omega_i$ is the characteristic frequency for atom $i$, defined as \cite{TS}:
\begin{equation}
\omega_i = \dfrac{4\ C_{6,i}^\text{free}}{3\ (\alpha_i^\text{0,free})^2}.
\end{equation}
The above ``effective'' terms are related to their free-atom values with proper scaling, i.e.,
\begin{equation}
    \alpha_{i}^\text{0,eff} = \alpha_{i}^\text{0,free} \left(\dfrac{V_i^{\text{eff}}}{V_i^{\text{free}}}\right),
\end{equation}
where the factor $V_i^{\text{eff}}/V_i^{\text{free}}$ represents the ratio of the effective volume occupied by an atom when interacting with its environment, as determined through the Hirshfeld partitioning scheme \cite{Hirshfeld}, to the reference volume of the same atom in a free (non-interacting) state. Generally, these volume ratios vary across an atomic structure and need to be retrieved from high fidelity models for accurate representation. In this work, however, we assume the ratio to be a constant value, as its variation is less than 1\% in the polymer melts under consideration. The MBD energy is finally evaluated as the difference between interacting and non-interacting frequencies as presented in Eq.~\ref{eq:E_mbd} and the MBD force is formulated as in Eq.~\ref{eq:F_mbd}, both in Section~\ref{sec:mbd}.

 \bibliographystyle{elsarticle-num}
 \bibliography{cas-refs}






\end{document}